\documentclass[10pt,twocolumn,letterpaper]{article}

\usepackage[pagenumbers]{cvpr} 

\usepackage[accsupp]{axessibility} 
\usepackage[dvipsnames]{xcolor}

\usepackage{colortbl}
\usepackage{xcolor}
\usepackage[T1]{fontenc}
\usepackage[latin9]{inputenc}
\usepackage{array}
\usepackage{multirow}
\PassOptionsToPackage{normalem}{ulem}
\usepackage{ulem}
\usepackage{arydshln}
\usepackage{array}
\usepackage{marvosym}
\providecommand{\tabularnewline}{\\}

\definecolor{cvprblue}{rgb}{0.21,0.49,0.74}
\usepackage[pagebackref,breaklinks,colorlinks,citecolor=cvprblue]{hyperref}

\def\paperID{4462} 
\def\confName{CVPR}
\def\confYear{2024}

\title{MemFlow: Optical Flow Estimation and Prediction with Memory}

\author{Qiaole Dong and Yanwei Fu\textsuperscript{\Letter}\\
School of Data Science, Fudan University\\
{\tt\small \{qldong18, yanweifu\}@fudan.edu.cn}
}

\begin{document}
\maketitle
\begin{abstract}
Optical flow is a classical task that is important to the vision community. Classical optical flow estimation uses two frames as input, whilst some recent methods consider multiple frames to explicitly model long-range information. The former ones limit their ability to fully leverage temporal coherence along the video sequence; and the latter ones incur heavy computational overhead, typically not possible for real-time flow estimation. Some multi-frame-based approaches even necessitate unseen future frames for current estimation, compromising real-time applicability in safety-critical scenarios. To this end, we present MemFlow, a real-time method for optical flow estimation and prediction with memory. Our method enables memory read-out and update modules for aggregating historical motion information in real-time.
Furthermore, we integrate resolution-adaptive re-scaling to accommodate diverse video resolutions. Besides, our approach seamlessly extends to the future prediction of optical flow based on past observations. Leveraging effective historical motion aggregation, our method outperforms VideoFlow with fewer parameters and faster inference speed on Sintel and KITTI-15 datasets in terms of generalization performance. 
At the time of submission, MemFlow also leads in performance on the 1080p Spring dataset. Codes and models will be available at: \url{https://dqiaole.github.io/MemFlow/}.

\end{abstract}

\section{Introduction}
\label{sec:intro}

\begin{figure}
\centering
\includegraphics[width=0.9\linewidth]{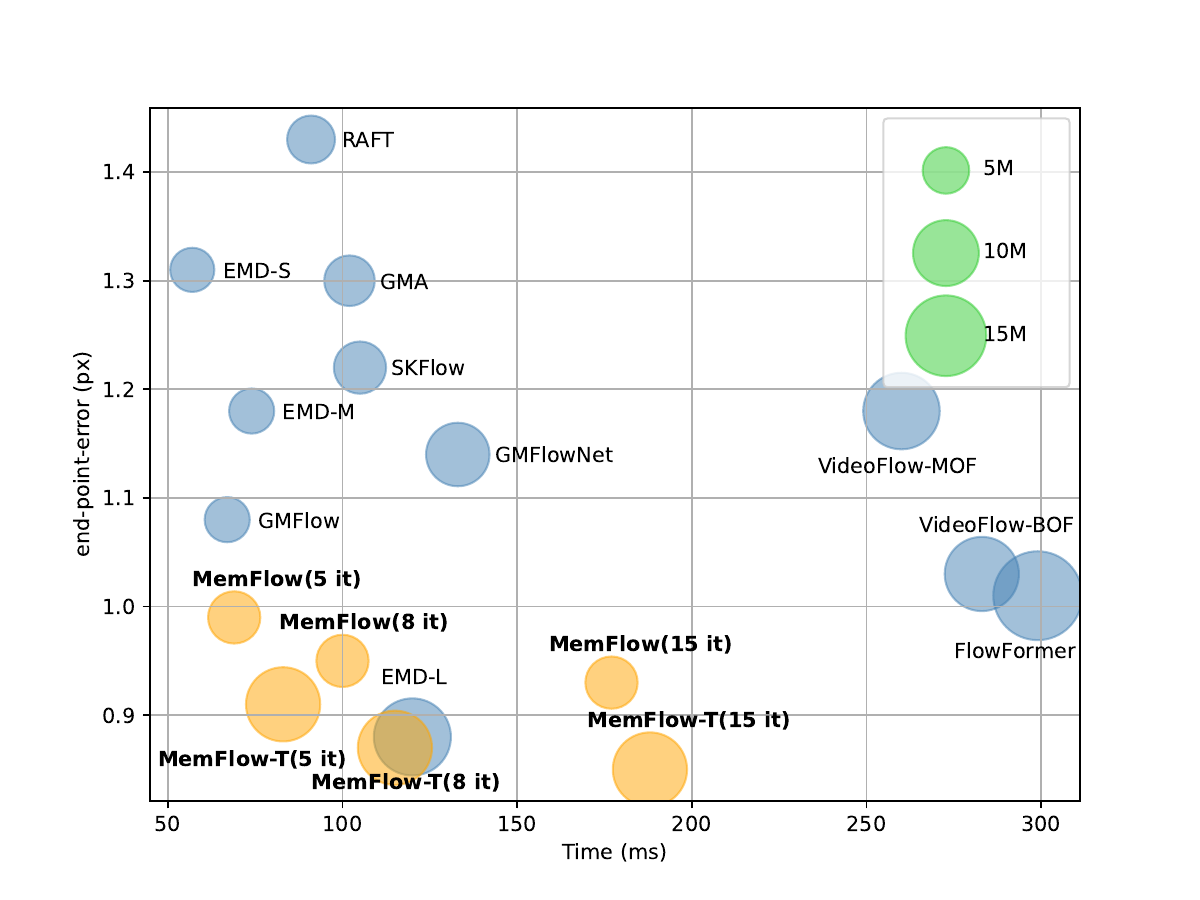}
\vspace{-0.15in}
\caption{End-point-error on Sintel (clean) vs. inference time (ms) and model size (M). All models are trained on FlyingChairs and FlyingThings3D, and tested with one NVIDIA A100 GPU. MemFlow(-T) (x it) indicates running our network with only x iterations of GRU. Our MemFlow(-T) achieves significant reductions in computational overhead as well as substantial performance boosts over the state-of-the-art methods.\label{fig:epe_time_para}}
\vspace{-0.15in}
\end{figure}


Optical flow, a critical area in computer vision, plays a key role in various real-world applications like video inpainting~\cite{gao2020flow}, action recognition~\cite{sun2018optical}, and video prediction~\cite{wu2022optimizing, geng2022comparing}.  In essence, it captures the displacement vector field for each pixel between successive video frames. Recent advances in optical flow estimation, as highlighted by works such as FlowNet~\cite{ilg2017flownet}, PWC-Net~\cite{sun2018pwc}, RAFT~\cite{teed2020raft}, SKFlow~\cite{sun2022skflow}, FlowFormer~\cite{huang2022flowformer}, and 
a rethinking training approach by MatchFlow\cite{dong2023rethinking}, 
have been successful. This success is attributed to advancements in model architectures~\cite{sun2018pwc, teed2020raft,huang2022flowformer} and dedicated datasets~\cite{dosovitskiy2015flownet, mayer2016large, dong2023rethinking}.

The classical optical flow works use two frames as input, potentially limiting their ability to fully leverage temporal coherence along a video sequence. This limitation results in a bottleneck, prompting an increasing reliance on computationally intensive vision transformer encoders~\cite{Chu2021twins} for improved performance, as noted in~\cite{huang2022flowformer}.

Conversely, some recent approaches~\cite{ren2019fusion, lu2023transflow, shi2023videoflow} explore the use of multi-frame videos as input. Typically, these methods either employ simple fusion modules with modest improvements or explicitly model long-range information, incurring heavy computational overhead.
For instance, PWC-Fusion~\cite{ren2019fusion} straightforwardly fuses backward warped past flow with current flow, resulting in a modest improvement of 0.65\% over the baseline PWC-Net~\cite{sun2018pwc}. TransFlow~\cite{lu2023transflow} and VideoFlow~\cite{shi2023videoflow} explicitly model long-range motion within a 5-frame context, leading to a significant computational overhead. Importantly, these methods~\cite{lu2023transflow,shi2023videoflow} operate in an offline mode, demanding access to unseen future frames in advance for current estimation. Additionally, VideoFlow runs considerably slower than its 2-frame baseline, SKFlow~\cite{sun2022skflow}, as depicted in \cref{fig:epe_time_para}. The substantial number of parameters (13.5M) also poses a significant burden on model deployment, causing out-of-memory issues when tested on the 1080p Spring dataset~\cite{mehl2023spring} with a single NVIDIA A100 80 GB GPU.

To this end,  we present \textbf{MemFlow}, an innovative architecture by proposing the memory module~\cite{oh2019video, wu2022memvit, cheng2021rethinking, hu2021learning, yu2022memory, li2022recurrent, lee2021video, wang2021temporal, paul2021local, cheng2022xmem} for effective optical flow estimation. It operates in real-time (online mode) efficiently with the following notable strengths: (1) \textit{Strong Cross-dataset Generalization Performance}. On both  clean and final pass of the Sintel~\cite{butler2012naturalistic} dataset, our model achieves an end-point-error (EPE) of 0.93 and 2.08 pixels. This represents a substantial 23.8\% and 15.4\% error reduction compared to our 2-frame baseline, SKFlow~\cite{sun2022skflow} (1.22 and 2.46 pixels).
When evaluated on the KITTI dataset~\cite{geiger2012we}, our method demonstrates an error rate of 13.7\%, showcasing an 11.6\% improvement over SKFlow (15.5\%).
%
(2) \textit{High Inference Efficiency}. Our MemFlow achieves an impressive inference speed of 5.6 frames per second (fps) on A100 GPU for processing 1024x436 videos.  Even a faster variant of our model, with 5 iterations of GRU, can run at 14.5 fps. Importantly,  this accelerated version maintains the near-best generalization performance, as illustrated  in \cref{fig:epe_time_para}.


Technically, our MemFlow is a real-time approach designed for optical flow estimation and prediction, incorporating a memory component. Specifically, MemFlow maintains a memory buffer that stores both historical motion information and context features from the input video stream. As new frame pairs are inputted, the memory buffer is continually updated with the latest context and motion features. We employ an attention mechanism to query the memory buffer using the context feature, extracting useful motion information as the aggregated motion feature. By combining this aggregated motion feature with the current motion and context features, we can regress the residual flow. 

Additionally, we introduce a resolution-adaptive re-scaling for similarity computation within the attention mechanism to enhance cross-resolution generalization during inference. Furthermore, we provide the option to replace our feature encoder with a more robust vision transformer~\cite{Chu2021twins}, referred to as \textbf{MemFlow-T}, resulting in improved outcomes.
As depicted in \cref{fig:epe_time_para}, our MemFlow (-T) demonstrates significant reductions in computational overhead and substantial performance enhancements compared to state-of-the-art methods. Notably, even with only 5 iterations, our MemFlow surpasses the performance of heavyweight state-of-the-art approaches such as VideoFlow~\cite{shi2023videoflow} and FlowFormer~\cite{huang2022flowformer}, while running faster than RAFT~\cite{teed2020raft}.


In addition to estimating optical flow, we're investigating future flow prediction using our versatile memory module.  This capability is crucial for intelligent systems like autonomous vehicles and robots, enabling effective planning and response in dynamic environments. Our adapted network for flow prediction is named \textbf{MemFlow-P}. 
In MemFlow-P, we maintain a memory buffer for decoding residual flow from context and aggregated motion features, eliminating the need for 2D motion features between the current and next frame. Upon the arrival of the next frame, we update the memory module with a new motion feature calculated between the current and next frames.
To showcase our flow prediction quality, we combine MemFlow-P with Softmax Splatting~\cite{Niklaus_CVPR_2020} and image inpainting~\cite{dong2022incremental, cao2023zits++, cao2022learning} for video prediction, involving the synthesis of future video frames based on past ones.
Despite not being specifically trained for video prediction, our method achieves comparable results with two competitive flow-based methods~\cite{wu2022optimizing, geng2022comparing} in SSIM~\cite{wang2004image} and LPIPS~\cite{zhang2018unreasonable}.


In summary, we make four significant contributions:
1) \textit{Innovative Real-Time Optical Flow Estimation}: We introduce a novel architecture that effectively employs a memory module, allowing for real-time optical flow estimation.
2) \textit{Enhanced Generalization with Resolution-Adaptive Re-scaling}: We propose the use of resolution-adaptive re-scaling in attention computation, enhancing cross-resolution generalization performance.
3) \textit{Superior Optical Flow Estimation}: Our MemFlow(-T) achieves state-of-the-art or near-SOTA performance on various standard optical flow estimation benchmarks, demonstrating exceptional performance with minimal computational overhead.
4) \textit{Future Prediction Capability without Explicit Training}: Repurposing MemFlow for optical flow future prediction, we achieve competitive results in video prediction without the need for specific training for this downstream task, highlighting the adaptability of our approach.

\section{Related Work}
\label{sec:related_work}

\begin{figure*}
\centering
\includegraphics[width=0.8\linewidth]{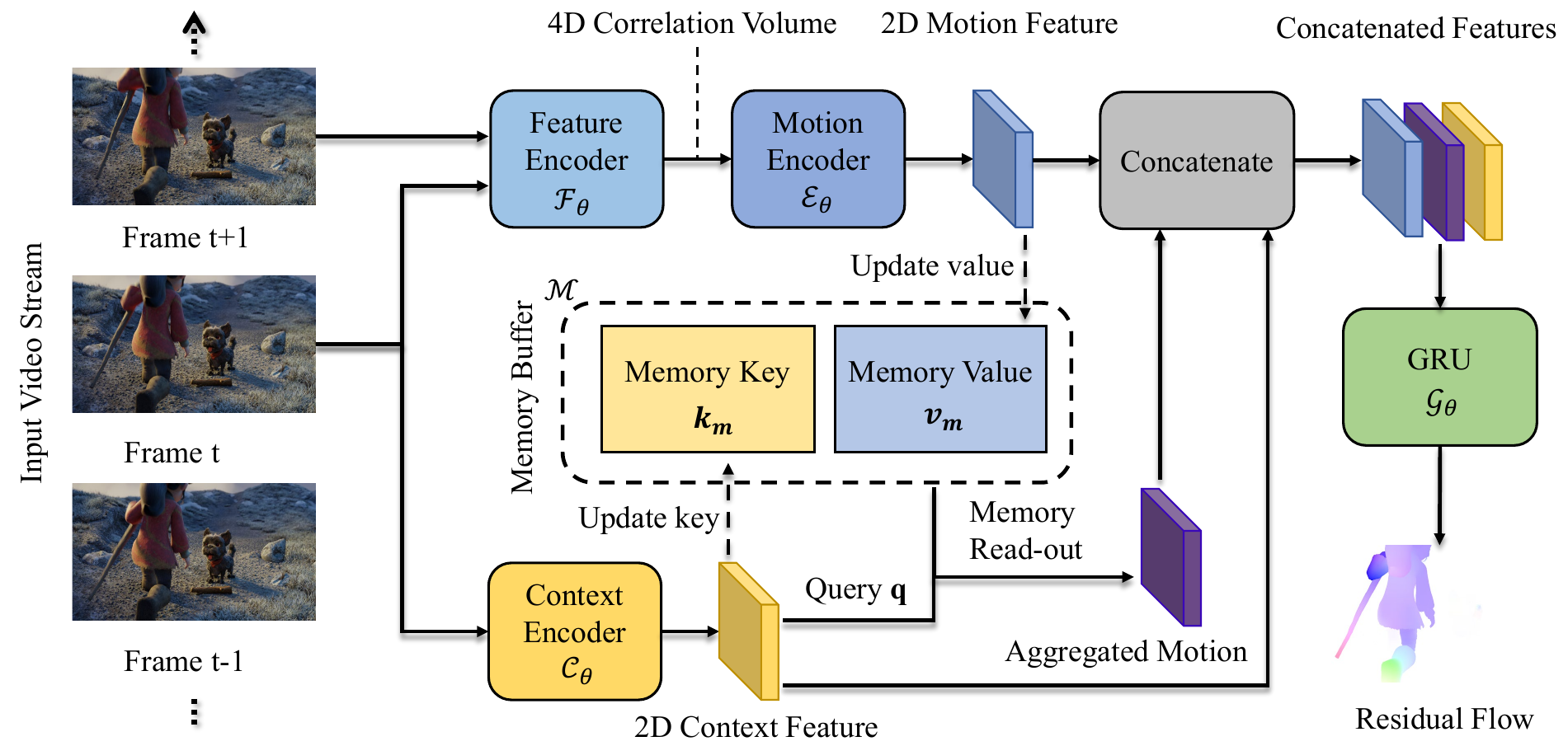}
\vspace{-0.15in}
\caption{Overview of our MemFlow. MemFlow maintains a memory buffer to store historical motion states of video, together with an efficient update and read-out process that retrieves useful motion information for the current frame's optical flow estimation. It has three key components: 1) \textit{Feature Extractors}. Feature and motion encoder extract and construct the motion feature for the current frame. Another context encoder produces the context feature. 2) \textit{Memory buffer}. Memory buffer stores historical context and motion features and read-out the aggregated motion feature.  3) \textit{Update Modules}. GRU updates the optical flow with a series of residual flows. And the Memory buffer is kept updating when a new frame comes.\label{fig:overview}}
\vspace{-0.15in}
\end{figure*}

\noindent\textbf{Optical Flow Estimation by Two Frames}. Traditionally, it is  solved by  optimizing energy function~\cite{horn1981determining, black1993framework, black1996robust, zach2007duality, revaud2015epicflow, brox2004high, bruhn2005lucas} to maximize visual similarity between images. Recent efforts resort to  
deep neural networks for directly regressing~\cite{dosovitskiy2015flownet, ilg2017flownet, sun2018pwc, hui2018liteflownet, ranjan2017optical, deng2023explicit} or generating~\cite{saxena2024surprising, dong2023open} optical flow from two frames. Specifically, FlowNet~\cite{dosovitskiy2015flownet} first proposed optical flow estimation with end-to-end trainable CNN. PWC-Net~\cite{sun2018pwc} and Lite-FlowNet~\cite{hui2018liteflownet} modified the network following the strategy of coarse-to-fine. RAFT~\cite{teed2020raft} further developed a convolutional GRU block upon multi-scale 4D correlation volume for iteratively updating. The following works~\cite{jiang2021learning,zheng2022dip,sun2022skflow,huang2022flowformer, dong2023rethinking, lu2023transflow, shi2023videoflow} continually improve the performance of optical flow estimation based on this recurrent architecture. 
Our MemFlow is also built upon and benefited from the most recently developed GRU-based network, while we employ memory to maintain past motion information and attention for gathering temporal cues for optical flow estimation.


\noindent\textbf{Optical Flow Estimation by Multiple Frames}. Traditional works~\cite{elad1998recursive, chin1994probabilistic} employed Kalman filter for optical flow estimation with temporal coherence. Some recent un/self-supervised deep models~\cite{janai2018unsupervised, liu2019selflow, liu2020learning} take three frames to estimate the optical flow of the current frame. 
On the other hand, there are also several supervised efforts. 
Particularly, PWC-Fusion~\cite{ren2019fusion} fused the backward warped past flow with current flow through a fusion module. 
RAFT~\cite{teed2020raft} implicitly utilized the historical frames as the initialization to  "warm-start" the optical flow estimation of current frame. TransFlow~\cite{lu2023transflow} and VideoFlow~\cite{shi2023videoflow} take five frames (including both past and future frames) to better model the long-range temporal information, while this demands prohibitive computational cost and memory footprint to store and process these frames. 
%
%
%
To make a balance of performance and cost, our MemFlow integrates a memory module using past frames
for optical flow, and thus is capable of being running in an online mode.
Moreover, MemFlow outperforms these competitors with better generalization performance while being much more computationally friendly.


\noindent\textbf{Future Prediction by Flows}. It aims to predict the optical flow into the future based on past frames~\cite{luo2017unsupervised, jin2017predicting}, motion history~\cite{ciamarra2022forecasting} or even single image~\cite{walker2015dense, walker2016uncertain, argaw2021optical}. 
Luo~\etal~\cite{luo2017unsupervised} firstly proposed to predict future 3D flow through a convolutional LSTM architecture. And OFNet~\cite{ciamarra2022forecasting} employed a UNet and ConvLSTM to predict the optical flow autoregressively based on past flows. However, Walker~\etal~\cite{walker2015dense, walker2016uncertain} predicted the optical flow from a single image and utilized variational autoencoders to model the uncertainty. In contrast, we repurpose our MemFlow for one time step ahead of optical flow prediction with minimal changes and achieve better prediction performance compared to recent OFNet~\cite{ciamarra2022forecasting} and several strongly competitive baselines.

\section{MemFlow}
\label{sec:Methodology}

\subsection{Definition and Overview}

\noindent\textbf{Problem Setup}. 
Optical flow is a per-pixel displacement vector field: $\mathbf{f}_{t\rightarrow t+1}=(\mathbf{f}^1, \mathbf{f}^2)$, mapping the location $(u, v)$ in current frame $\mathbf{I}_t$ to next frame $\mathbf{I}_{t+1}$ as $(u+f^1(u), v+f^2(v))$. In the setting of our online multi-frame optical flow estimation, we have access to memory of past history and current frame pair: $\mathbf{I}_t, \mathbf{I}_{t+1}$. We aim to output an optical flow $\mathbf{f}_{t\rightarrow t+1}$ and update the memory accordingly. Note that the same MemFlow framework can be directly utilized to estimate optical flow for future prediction. Thus such a task is also evaluated here: we only get video frame $\mathbf{I}_t$ and memory of past, while predicting optical flow $\mathbf{f}_{t\rightarrow t+1}$ into future. 

\noindent\textbf{Overview.}  We present an overview of our Memory module for optical Flow estimation (MemFlow) as in \cref{fig:overview}. 
Specifically, our MemFlow consists of three key components: 1) \textit{Feature Extractors}. We have a feature encoder $\mathcal{F}_\theta$ extracting features from the frames to construct the 4D correlation volume subsequently. By correlation lookup operation~\cite{teed2020raft}, the following motion encoder $\mathcal{E}_\theta$ can produce the motion feature. To provide 
context features of the current frame, we also employ the context encoder $\mathcal{C}_\theta$.
%
%
2) \textit{Memory buffer}. We store historical context and motion features in the buffer $\mathcal{M}$, while only the aggregated motion feature can be read-out through an attention mechanism. 3) \textit{Update Modules}. We iterate the modules of GRU and Memory for flow and feature updating, respectively. Typically,
GRU $\mathcal{G}_\theta$ outputs a series of residual flows. And we update the memory buffer with the final optical flow.

In the next sections, we'll elaborate on our proposed memory module for optical flow estimation (MemFlow(-T)) in \cref{subsec:memflow_ofe,subsec:mem_update}. Subsequently, we'll demonstrate its application in future prediction through minimal modifications, resulting in our optical flow prediction model, MemFlow-P, as outlined in \cref{subsec:memflow_p_ofp}.


\subsection{Memory Read-out}
\label{subsec:memflow_ofe}

We will first introduce the necessary feature extraction module, then present our novel memory read-out and resolution-adaptive re-scaling for the aggregated motion feature here.

\noindent \textbf{Feature Extraction}. 
Given current input image pairs: $\mathbf{I}_{t}, \mathbf{I}_{t+1}$, we first extract the feature of images at 1/8 resolution by feature encoder $\mathcal{F}_\theta$ : $\mathcal{F}_\theta(\mathbf{I}_{t}), \mathcal{F}_\theta(\mathbf{I}_{t+1})\in \mathbb{R}^{H\times W\times D}$, where $D$ is the number of channel; $H, W$ indicate the 1/8 height and width of original images. We then construct the 4D correlation volume $C$ through the dot product between all pairs of features:
\begin{equation}
 C =  \mathcal{F}_\theta(\mathbf{I}_{t})\times \mathcal{F}_\theta(\mathbf{I}_{t+1})^T\in\mathbb{R}^{H\times W\times H\times W}.
\end{equation}
%
With the current estimation of optical flow $f_i$, which is initialized as an all zeros tensor, we can lookup correlation values from $C$ as in \cite{teed2020raft}. Combined with the current flow, we can get the motion feature  $f_{m}=\mathcal{E}_\theta(f_i, \mathrm{LookUp}(C, f_i))$.
Finally, we extract the context feature $f_{c}$ from $\mathbf{I}_{t}$ with our context encoder $\mathcal{C}_\theta$, which is trained with the same network architecture of feature encoder $\mathcal{F}_\theta$.
Please refer to the implementation and supplementary for the network details.

\noindent \textbf{Memory Read-out}. 
We present a novel module for memory read-out. 
Our memory buffer $\mathcal{M}=\{k_m\in\mathbb{R}^{L\times D_k}, v_m\in\mathbb{R}^{L\times D_v}\}$, initialized from an empty set, consists of memory keys and values, where $L=l\times H\times W$ is the number of keys and values. $D_k, D_v$ are the feature dimension. We further define $l$ as the length of memory buffers. With current context feature $f_{c}$ and motion feature $f_{m}$, we read-out the aggregated motion feature through an attention mechanism. Specifically, we first linear project $f_{c}, f_{m}$ and get the corresponding query, key, and value by concatenation with memory buffer,
\begin{equation}\label{equ:mem_concat}
q=f_{c}W_q, \quad k=[f_{c}W_k; k_m],\quad v=[f_{m}W_v; v_m],
\end{equation}
where $W_q, W_k, W_v$ are the learnable projection parameters, and $[;]$ is the concatenation operation along first dimension. The aggregated motion feature can be read-out by
\begin{equation}\label{equ:mem_read_out}
f_{am} = f_m + \alpha\cdot \mathrm{Softmax}(1/\sqrt{D_k} \times q\times k^T)\times v,
\end{equation}
where $\alpha$ is a learnable scalar initialized from 0. And we omit the necessary reshape operation here for simplicity.
Note that, GMA~\cite{jiang2021learning} utilizes attention to aggregate spatial information, while we employ the attention for gathering additional temporal information, as illustrated in \cref{equ:mem_concat}. Furthermore, we enhance the attention for resolution adapting through a re-scaling technique, as explained later.

\noindent\textbf{Resolution-adaptive Re-scaling.} We also introduce a novel strategy for adapting resolution here. Specifically, 
%
if the model is trained using sequences up to length $N$, it struggles to generalize attention effectively to sequences longer than $N$.
Pioneer work~\cite{chiang2022overcoming} found that the dilution of similarity score accounts for this. So Chiang and Cholak~\cite{chiang2022overcoming} proposed to fix this problem by scaling similarity with $\log n$, where $n$ is the sequence length. In contrast to them, we further update the scaling with
average training sequence length $n_{avg}$ as the logarithmic base, and use the length of key $k$ as the sequence length in our cross-attention. 
So the softmax function in \cref{equ:mem_read_out} is updated as
%
\begin{equation}
\mathrm{Softmax}(\frac{\log_{n_{avg}}(L+H*W)}{\sqrt{D_k}} \times q\times k^T).
\end{equation}
After incorporating this novel scaling coefficient into memory read-out, it can work for various resolutions and even generalize well to 1080p video as verified in the experiment.

\subsection{Memory Update and Flow Estimation}
\label{subsec:mem_update}
In this section, we introduce a novel memory update strategy and flow estimation with our new memory module.

Particularly, with the context, motion, and aggregated motion features, we can now output a residual flow through a GRU unit: $\Delta f_i=\mathrm{GRU}(f_c, f_{m}, f_{am})$. After N iterations of GRU, we can get the final optical flow and corresponding motion feature $f_m$. We then update the memory buffer by inserting the transformed context and motion feature into the key and value tensors of memory,
\begin{equation}
k_m=[f_{c}W_k; k_m],\quad v_m=[f_{m}W_v; v_m].
\end{equation}
When the memory buffer length $l$ exceeds a pre-defined maximum of $l_{max}$, we simply discard the obsolete features. Though we try to distill these obsolete features into long-term memory and model the long-range motion information, we find it has no effect on the final performance.

\noindent\textbf{Loss Functions.} Our loss functions are inherited from the classical works - RAFT~\cite{teed2020raft}. Generally,
we supervise our network with $l_1$ distance between our partially summed residual flow $\{f_1, \dots, f_N\}$ and groundtruth $f_{gt}$ with exponentially increasing weights,
\begin{equation}
\mathcal{L}=\sum_{i=1}^N0.85^{N-i}||f_{gt}-f_i||_1, \quad N=12.
\end{equation}

\subsection{Beyond Flow Estimation: Future Prediction}
\label{subsec:memflow_p_ofp}

The framework in \cref{fig:overview} can be directly utilized for future prediction with minimal changes, and we present the modifications here. More details are in the supplementary.

As we do not have access to frame $I_{t+1}$, we are not able to calculate the correlation volume and encode the motion feature for the current frame. So we extract the context feature $f_c$ from the current frame and read-out the aggregated motion feature $f_{am}$ from the memory buffer as in \cref{subsec:memflow_ofe}. Then we predict the optical flow by a small convolutional network based on $f_c$, $f_{am}$, and past flow $f_p$: $f=\mathrm{Convs}(f_c, f_{am}, f_p)$. After the next frame comes, we can now calculate motion feature based on our predicted flow or flow estimated by MemFlow. Finally, we could update the memory buffer of MemFlow-P as in \cref{subsec:mem_update} and are ready for optical flow prediction of the next frame. We also use $l_1$ distance as our loss function.

We further utilize the flow prediction for video prediction. Generally, we forward warp the last video frame by Softmax Splatting~\cite{Niklaus_CVPR_2020} with monocular depth from DPT~\cite{ranftl2021vision} and our predicted optical flow. And we will fill the holes due to splatting with image inpainting method~\cite{dong2022incremental}.


\section{Experiments}
\label{sec:Experiments} 

\noindent\textbf{Dataset and Implementation.} We adopt SKFlow~\cite{sun2022skflow} as the network architecture of MemFlow(-P). And we further replace the feature encoder of SKFlow with Twins-SVT~\cite{Chu2021twins} as MemFlow-T. The maximum length of memory buffer $l_{max}$ is set to 1. The iteration number of GRU is set to 15 by default during inference. In order to learn better correlation and motion features, we first pre-train our network with 2-frame input on FlyingChair~\cite{dosovitskiy2015flownet} and FlyingThings3D~\cite{mayer2016large} following SKFlow. Subsequently, with 3-frame video as input, we train MemFlow(-T) and MemFlow-P on FlyingThings3D for generalization evaluation and then finetune for Sintel~\cite{butler2012naturalistic} submission with the combination of Sintel, KITTI~\cite{geiger2012we}, HD1K~\cite{kondermann2016hci}, FlyingThings3D. Finally, we finetune the model with KITTI and newly proposed Spring~\cite{mehl2023spring} for KITTI and Spring submission, respectively. Our network is trained with AdamW~\cite{loshchilov2017decoupled} optimizer with one-cycle~\cite{smith2019super} learning rate on two NVIDIA A100 GPUs. Further details are provided in the supplementary material.

\noindent\textbf{Evaluation Metric.} We utilize end-point-error (EPE) and Fl-all for Sintel and KITTI evaluation. EPE denotes the $l_2$ distance between estimated flow and groundtruth. And Fl-all refers to the percentage of outliers whose EPE is larger than 3 pixels and 5\% of groundtruth flow magnitude. For the Spring benchmark, we also adopt 1-pixel outlier rate (1px) and WAUC~\cite{richter2017playing}, which is the weighted average of the inlier rates for a range of thresholds from 0 to 5 pixels. In the following tables, the best results are in bold, while the second-best ones are underlined.

\subsection{Optical Flow Estimation}
\begin{figure}
\centering
\includegraphics[width=0.6\linewidth]{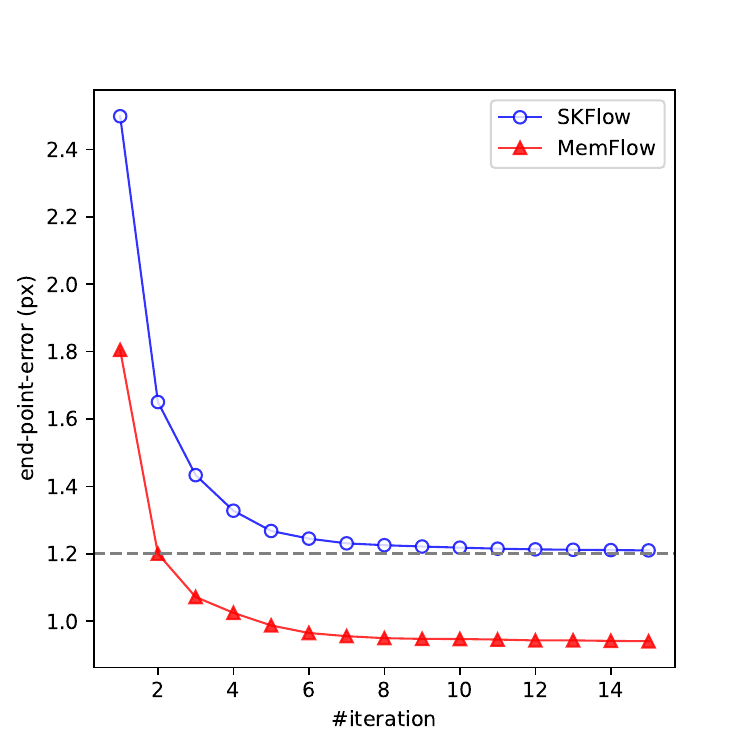}
\vspace{-0.15in}
\caption{End-point-error of optical flow vs. number of iterations during inference. This figure provides the generalization performance on Sintel (clean) training set. Our method outperforms 15-iteration SKFlow's performance, after using only 2 iterations.\label{fig:epe_iters}}
\vspace{-0.15in}
\end{figure}

\begin{figure*}
\centering
\includegraphics[width=0.85\linewidth]{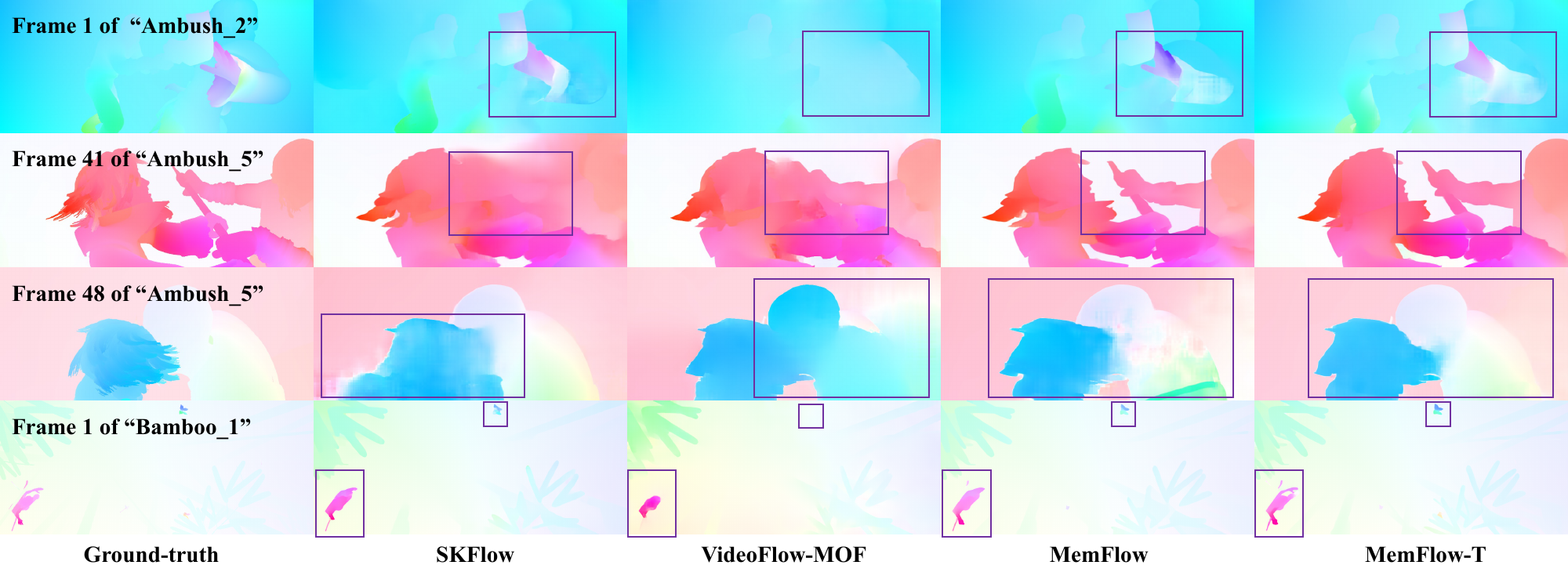}
\vspace{-0.15in}
\caption{Qualitative comparison on the training set of Sintel final pass after pre-training on FlyingChair and FlyingThings3D. Notable areas are marked by a bounding box. Please zoom in for details.\label{fig:sintel_qualitative}}
\vspace{-0.15in}
\end{figure*}

\noindent\textbf{Generalization Performance.} Following previous works, we first show the generalization performance as in \cref{tab:generalize_s_k}. Our MemFlow(-T) achieves state-of-the-art zero-shot performance on both challenging datasets, even with fewer iterations and inference time as shown in \cref{fig:epe_time_para}. MemFlow with 5 iterations of GRU can even run in real-time while still keeping near-SOTA in terms of generalization. Particularly, though share the same model architecture, our MemFlow reduces EPE by 0.38 and 0.39 from SKFlow~\cite{sun2022skflow} on Sintel final pass and KITTI-15, respectively. Besides, we visualize the EPE evolution over iterations in \cref{fig:epe_iters}. With only 2 iterations, our MemFlow can beat the previous SKFlow, showing superior efficiency. \cref{fig:sintel_qualitative} further provides a qualitative comparison on Sintel final pass with SKFlow and VideoFlow-MOF~\cite{shi2023videoflow}. Our MemFlow(-T) not only exhibits more accurate details on large motion regions of hands and head but also are good at fine detail of small object. 

Besides, we find that recent multi-frame based VideoFlow is typically not good at estimating optical flow for the first frame of a video. Because VideoFlow needs strict 3-frame or 5-frame input, and output optical flow for the center frame only. So they need to copy the first frame and insert it into the video as a pseudo previous frame, which accounts for why it tends to output some zero flow for the first frame as shown in the first and fourth row of \cref{fig:sintel_qualitative}. Yet our MemFlow can work normally without regard to the position of the input frame within the video.

\begin{table} \small
\centering
\caption{Generalization performance of optical flow estimation on Sintel and KITTI-15 after trained on FlyingChairs and FlyingThings3D. \emph{MF} indicates methods using multi frames for optical flow.\label{tab:generalize_s_k}}
\vspace{-0.1in}
\begin{tabular}{ccccc}
\toprule
\multirow{2}{*}{Model} & \multicolumn{2}{c}{Sintel} & \multicolumn{2}{c}{KITTI-15}\tabularnewline
\cline{2-5}
 & Clean & Final & Fl-epe & Fl-all\tabularnewline
\midrule
RAFT~\cite{teed2020raft} & 1.43 & 2.71 & 5.04 & 17.4\tabularnewline
GMA~\cite{jiang2021learning} & 1.30 & 2.74 & 4.69 & 17.1\tabularnewline
GMFlow~\cite{xu2022gmflow} & 1.08 & 2.48 & 7.77 & 23.4\tabularnewline
GMFlowNet~\cite{zhao2022global} & 1.14 & 2.71 & 4.24 & 15.4\tabularnewline
SKFlow~\cite{sun2022skflow} & 1.22 & 2.46 & 4.27 & 15.5\tabularnewline
MatchFlow~\cite{dong2023rethinking} & 1.03 & 2.45 & 4.08 & 15.6\tabularnewline
FlowFormer++~\cite{shi2023flowformer++} & 0.90 & 2.30 & 3.93 & 14.1\tabularnewline
EMD-L~\cite{deng2023explicit} & \uline{0.88} & 2.55 & 4.12 & \uline{13.5}\tabularnewline
\hdashline
TransFlow\textsuperscript{(\emph{MF})}~\cite{lu2023transflow} & 0.93 & 2.33 & 3.98 & 14.4\tabularnewline
VideoFlow-BOF\textsuperscript{(\emph{MF})}~\cite{shi2023videoflow} & 1.03 & 2.19 & 3.96 & 15.3\tabularnewline
VideoFlow-MOF\textsuperscript{(\emph{MF})}~\cite{shi2023videoflow} & 1.18 & 2.56 & 3.89 & 14.2\tabularnewline
\textbf{MemFlow (Ours)}\textsuperscript{(\emph{MF})}  & 0.93 & \uline{2.08} & \uline{3.88} & 13.7\tabularnewline
\textbf{MemFlow-T (Ours)}\textsuperscript{(\emph{MF})}  & \textbf{0.85} & \textbf{2.06} & \textbf{3.38} & \textbf{12.8}\tabularnewline
\bottomrule
\end{tabular}
\vspace{-0.1in}
\end{table}

\begin{figure*}
\centering
\includegraphics[width=0.85\linewidth]{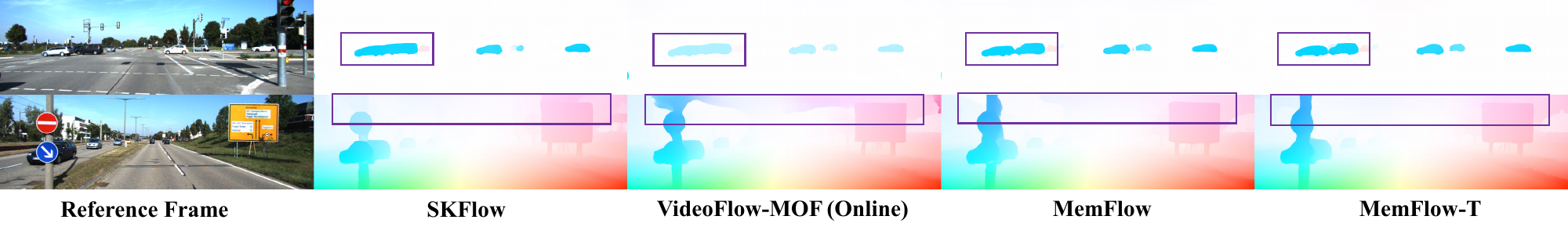}
\vspace{-0.15in}
\caption{Qualitative comparison on test set of KITTI-15 after finetuning. Ours do much better at distinguishing between different vehicles (first row) and between the foreground and the sky (second row). Please zoom in for details.\label{fig:kitti_qualitative}}
\vspace{-0.15in}
\end{figure*}

\noindent\textbf{Finetuning Evaluation.} We further report the finetuning results on public benchmark datasets, Sintel and KITTI, in \cref{tab:finetune_s_k}. Our MemFlow(-T) improves SKFlow by a large margin and outperforms most previous methods, \eg SOTA 2-frame based methods FlowFormer++~\cite{shi2023flowformer++} and multi-frame based TransFlow~\cite{lu2023transflow}, except recent VideoFlow. However, the online version of 5-frame VideoFlow-MOF degrades Fl-all on the KITTI-15 test set from 3.65 to 4.08, which is worse than our 3.88. We suspect that using multi-frame in an offline mode explicitly can lead to better dataset-specific performance, while with limited cross-dataset generalization performance as in \cref{tab:generalize_s_k}, where 5-frame VideoFlow-MOF performs much worse than 3-frame VideoFlow-BOF on Sintel. Finally, a qualitative result in \cref{fig:kitti_qualitative} on the KITTI test set shows our MemFlow(-T) outperforms others in distinguishing between different vehicles and between the foreground and the sky.

\begin{table} \small
\centering
\caption{Optical flow finetuning evaluation on the public benchmark. \emph{MF} indicates methods using multi frames for optical flow. * uses RAFT's multi-frame "warm-start" strategy on Sintel.\label{tab:finetune_s_k}}
\vspace{-0.1in}
\begin{tabular}{cccc}
\toprule
\multirow{2}{*}{Model} & \multicolumn{2}{c}{Sintel} & KITTI-15\tabularnewline
\cline{2-4}
 & Clean & Final & Fl-all\tabularnewline
\midrule
RAFT$^{*}$~\cite{teed2020raft} & 1.61 & 2.86 & 5.10\tabularnewline
GMA$^{*}$~\cite{jiang2021learning} & 1.39 & 2.47 & 5.15\tabularnewline
GMFlow~\cite{xu2022gmflow} & 1.74 & 2.90 & 9.32\tabularnewline
GMFlowNet~\cite{zhao2022global} & 1.39 & 2.65 & 4.79\tabularnewline
SKFlow$^{*}$~\cite{sun2022skflow} & 1.28 & 2.23 & 4.84\tabularnewline
MatchFlow$^{*}$~\cite{dong2023rethinking} & 1.16 & 2.37 & 4.63\tabularnewline
FlowFormer++~\cite{shi2023flowformer++} & 1.07 & 1.94 & 4.52\tabularnewline
EMD-L~\cite{deng2023explicit} & 1.32 & 2.51 & 4.51\tabularnewline
\hdashline
PWC-Fusion\textsuperscript{(\emph{MF})}~\cite{ren2019fusion} & 3.43 & 4.57 & 7.17\tabularnewline
TransFlow\textsuperscript{(\emph{MF})}~\cite{lu2023transflow} & 1.06 & 2.08 & 4.32\tabularnewline
VideoFlow-BOF\textsuperscript{(\emph{MF})}~\cite{shi2023videoflow} & \uline{1.01} & \uline{1.71} & 4.44\tabularnewline
VideoFlow-MOF\textsuperscript{(\emph{MF})}~\cite{shi2023videoflow} & \textbf{0.99} & \textbf{1.65} & \textbf{3.65}\tabularnewline
VideoFlow-MOF\textsuperscript{(\emph{MF})} (online)~\cite{shi2023videoflow} & - & - & 4.08\tabularnewline
\textbf{MemFlow (Ours)}\textsuperscript{(\emph{MF})}  & 1.05 & 1.91 & 4.10\tabularnewline
\textbf{MemFlow-T (Ours)}\textsuperscript{(\emph{MF})}  & 1.08 & 1.84 & \uline{3.88}\tabularnewline
\bottomrule
\end{tabular}
\vspace{-0.1in}
\end{table}

\begin{figure*}
\centering
\includegraphics[width=0.85\linewidth]{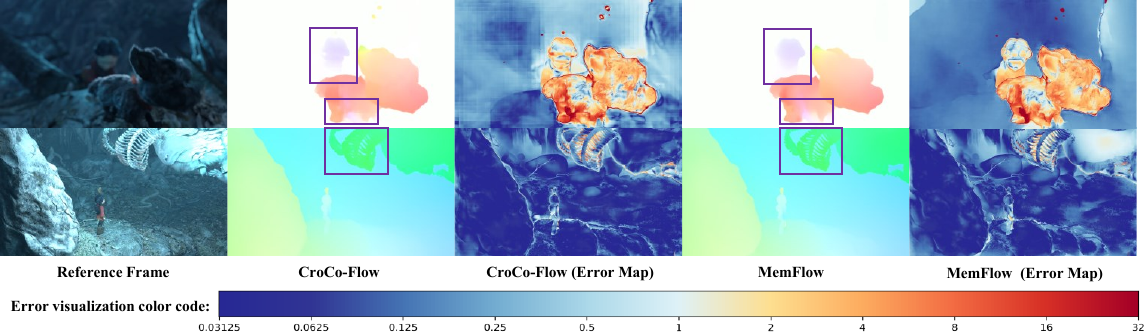}
\vspace{-0.1in}
\caption{Qualitative comparison with CroCo-Flow~\cite{weinzaepfel2023croco} on Spring test set. MemFlow performs better on fine details. Notable areas are also marked by a bounding box. Please zoom in for details.\label{fig:spring_qualitative}}
\vspace{-0.25in}
\end{figure*}

\noindent\textbf{Evaluation on Full-HD Spring Dataset.} We also report the generalization and finetuning results on the newly proposed Full-HD (1080p) Spring benchmark in \cref{tab:spring}. We first test MemFlow trained on Sintel for evaluation of generalization. \cref{tab:spring} shows that our MemFlow achieves the best EPE and Fl-all while being competitive with MS-RAFT+~\cite{jahedi2022high} in terms of 1px within different regions. After finetuning on the Spring training set, MemFlow outperforms previous SOTA CroCo-Flow~\cite{weinzaepfel2023croco} by a large margin, though CroCo-Flow is pretrained with additional 5.3M real-world image pairs. Qualitative comparison in \cref{fig:spring_qualitative} also shows that MemFlow performs better on fine details, while CroCo-Flow employs the much slower tile-technique~\cite{jaegle2021perceiver} for high-resolution testing and leads to block-like artifacts. Besides, we should point out that VideoFlow encounters out-of-memory when tested on the Spring dataset with a single NVIDIA A100 80 GB GPU. This further shows that our MemFlow is much more computationally friendly.

\begin{table*} \small
\setlength{\tabcolsep}{2pt}
\centering
\caption{Optical flow generalization and finetuning results on Spring~\cite{mehl2023spring}. We provide the 1px outlier rate for low/high-detail, (un)matched, (non-)rigid, and (not) sky regions. We also show the EPE, Fl error~\cite{geiger2012we}, and WAUC~\cite{richter2017playing}. Important metrics are highlighted in blue.\label{tab:spring}}
\vspace{-0.1in}
\begin{tabular}{cc >{\columncolor{blue!15}}c ccccccccccc >{\columncolor{blue!15}}c >{\columncolor{blue!15}}c >{\columncolor{blue!15}}c}
\toprule
\multirow{2}{*}{Dataset} & \multirow{2}{*}{Model} & \multicolumn{12}{c}{1px} &  &  & \tabularnewline
\cline{3-14}
 &  & total & low-det. & high-det. & matched & unmat. & rigid & non-rig. & not sky & sky & s0-10 & s10-40 & s40+ & \multirow{-2}{*}{EPE} & \multirow{-2}{*}{Fl} & \multirow{-2}{*}{WAUC}\tabularnewline
\midrule
\multirow{6}{*}{\rotatebox{90}{C+T+S+K+H}} & RAFT~\cite{teed2020raft} & 6.79 & 6.43 & 64.09 & 6.00 & 39.48 & 4.11 & 27.09 & 5.25 & 30.18 & 3.13 & 5.30 & 41.40 & 1.476 & 3.20 & 90.92\tabularnewline
 & GMA~\cite{jiang2021learning} & 7.07 & 6.70 & 66.20 & 6.28 & 39.89 & 4.28 & 28.25 & 5.61 & 29.26 & 3.65 & 5.39 & 40.33 & 0.914 & 3.08 & 90.72\tabularnewline
 & GMFlow~\cite{xu2022gmflow} & 10.36 & 9.93 & 76.61 & 9.06 & 63.95 & 6.80 & 37.26 & 8.95 & 31.68 & 5.41 & 9.90 & 52.94 & 0.945 & 2.95 & 82.34\tabularnewline
 & FlowFormer~\cite{huang2022flowformer} & 6.51 & 6.14 & 64.22 & 5.77 & 37.29 & 3.53 & 29.08 & 5.50 & \uline{21.86} & 3.38 & 5.53 & 35.34 & 0.723 & 2.38 & 91.68\tabularnewline
 & MS-RAFT+~\cite{jahedi2022high} & \textbf{5.72} & \textbf{5.37} & \textbf{61.50} & \textbf{5.04} & \uline{33.95} & \textbf{3.05} & \uline{25.97} & \uline{4.84} & \textbf{19.15} & \textbf{2.06} & \uline{5.02} & \uline{38.32} & \uline{0.643} & \uline{2.19} & \textbf{92.89}\tabularnewline
 & \textbf{MemFlow} & \uline{5.76} & \uline{5.39} & \uline{63.35} & \uline{5.11} & \textbf{32.76} & \uline{3.29} & \textbf{24.42} & \textbf{4.49} & 24.99 & \uline{2.92} & \textbf{4.82} & \textbf{32.07} & \textbf{0.627} & \textbf{2.11} & \uline{92.25}\tabularnewline
\hline
\multirow{2}{*}{+Spring} & CroCo-Flow~\cite{weinzaepfel2023croco} & 4.57 & 4.21 & \textbf{60.59} & 3.85 & \textbf{34.20} & \textbf{2.19} & 22.50 & 4.48 & \textbf{5.87} & \textbf{1.23} & \textbf{4.33} & 33.13 & 0.498 & 1.51 & 93.66\tabularnewline
 & \textbf{MemFlow} & \textbf{4.48} & \textbf{4.12} & 61.70 & \textbf{3.74} & 35.12 & 2.39 & \textbf{20.31} & \textbf{3.93} & 12.81 & 1.31 & 4.44 & \textbf{31.18} & \textbf{0.471} & \textbf{1.42} & \textbf{93.86}\tabularnewline
\bottomrule
\end{tabular}
\vspace{-0.1in}
\end{table*}

\subsection{Future Prediction of Optical Flow}
For future prediction of optical flow, we compare with following three baselines: (1) \emph{MemFlow}, we use MemFlow to estimate $\mathbf{f}_{t-1\rightarrow t}$ with available frames and forward warp the flow to next time step as $\hat{\mathbf{f}}_{t\rightarrow t+1}$. (2) \emph{Warped Oracle}, in contrast to (1), we forward warp the available optical flow groundtruth in dataset as a performance upper bound of the warping-based method. (3) \emph{OFNet}~\cite{ciamarra2022forecasting}, a learning-based method that trains a UNet and ConvLSTM with 6 past optical flows as input for future prediction. Note that all trainable models are trained on FlyingThings3D for comparison.

\noindent\textbf{Flow Prediction Results.} Left part of \cref{tab:of_prediction} reports the EPE on test split of FlyingThings3D, training set of Sintel and KITTI-15. MemFlow-P outperforms other competitors on all three datasets by a large margin, showing great dataset-specific and cross-dataset performance. More results can be found in supplementary material.

\noindent\textbf{Downstream Task: Video Prediction.} 
We show here quantitatively that our MemFlow-P generalizes well to video prediction. 
We compare our method with recent two flow-based video prediction models~\cite{wu2022optimizing, geng2022comparing} on KITTI. Though without training for video prediction specifically, our method can indeed achieve comparable or even better SSIM~\cite{wang2004image} and LPIPS~\cite{zhang2018unreasonable} as in the right part of \cref{tab:of_prediction}.

\begin{table}
 \small
\centering
\caption{\emph{Left}: End-point-error of flow prediction on FlyingThings3D (Final), Sintel (Final), and KITTI-15. \emph{Right}: Comparison of next frame prediction on KITTI test set (256x832). Note that our method is not trained for video prediction specifically.\label{tab:of_prediction}}
\vspace{-0.1in}
    \begin{tabular}{m{0.25\textwidth}m{0.25\textwidth}}
    \setlength{\tabcolsep}{1pt}
    \begin{tabular}{cccc}
\toprule
Method & Things & Sintel & KITTI\tabularnewline
\midrule
Warped Oracle & 14.76 & 5.76 & -\tabularnewline
\textbf{MemFlow} & 15.70 & 6.23 & 12.95\tabularnewline
\hdashline
OFNet~\cite{ciamarra2022forecasting} & 13.76 & 6.03 & 12.43 \tabularnewline
\textbf{MemFlow-P} & \textbf{7.56} & \textbf{5.38} & \textbf{8.82}\tabularnewline
\bottomrule
\end{tabular}
        &
        \setlength{\tabcolsep}{1pt}
        \begin{tabular}{ccc}
\toprule
Method & SSIM$\uparrow$ & LPIPS$\downarrow$\tabularnewline
\midrule
VPVFI~\cite{wu2022optimizing} & \textbf{0.827} & \textbf{0.123}\tabularnewline
VPCL~\cite{geng2022comparing} & 0.820 & 0.172\tabularnewline
\textbf{Ours} & \uline{0.825} & \uline{0.138}\tabularnewline
\bottomrule
\end{tabular}
        \\
    \end{tabular}
    \vspace{-0.2in}
\end{table}

\subsection{Ablation study}

In this section, we provide ablation studies on MemFlow by evaluating the generalization performance. We also finetune SKFlow with the same data volume as ours, denoted as SKFlow* in \cref{tab:abla_s_k}. Note that a $T$-frame video has $T-1$ flow labels. So the training step varies for fair comparison.

\noindent\textbf{2-Frame Pretraining.} We first assess the effect of 2-frame pre-training. As in \cref{tab:abla_s_k}, pretraining substantially improves the generalization performance on Sintel and KITTI-15, except for the slightly worse EPE of Sintel final pass.

\noindent\textbf{Inference Memory Length.} In this experiment, MemFlow is trained using 5-frame video on FlyingThings3D, and the maximum length of memory buffer $l_{max}$ is set to 2 at training. However, \cref{tab:abla_s_k} shows that during inference, set $l_{max}$ to 1 can achieve the best performance on three metrics. Compared to not using memory module during inference, enabling the memory module can improve the performance a lot, which shows the benefit of our designed memory module. But increasing $l_{max}$ from 1 to 3 can degrade the results. We think that it's because motion state a few frames ago is less relevant to current motion.

\noindent\textbf{Training Video Length.} We are also interested in how long the video we need for training the memory module. Fortunately, \cref{tab:abla_s_k} shows that video of 3-frame is good enough to train MemFlow. Training on longer video only results in comparable results but with much more computational overhead, \eg GPU memory and training time.

\noindent\textbf{Warm-start.} Warm-start was originally proposed by RAFT for better initialization with the previous flow. It's also compatible with MemFlow. However, equipped with our memory module, warm-start has little effect on the results as shown in \cref{tab:abla_s_k}. So we don't use warm-start by default.

\begin{table} \small
\setlength{\tabcolsep}{4pt}
\centering
\caption{Ablation studies. Parameters used in our final model are underlined. * means finetuning SKFlow~\cite{sun2022skflow} with  1200k steps.\label{tab:abla_s_k}}
\vspace{-0.1in}
\begin{tabular}{cccccc}
\toprule
\multirow{2}{*}{Experiment} & \multirow{2}{*}{Method} & \multicolumn{2}{c}{Sintel} & \multicolumn{2}{c}{KITTI-15}\tabularnewline
\cline{3-6}
 &  & Clean & Final & Fl-epe & Fl-all\tabularnewline
\midrule
Baseline & SKFlow$^{*}$ & 1.13 & 2.39 & 4.03 & 14.63\tabularnewline
\hline
\tabularnewline
\multicolumn{6}{c}{\noindent\emph{Reference Model, Training: 300k on FlyingThings, max Mem is 2}}\tabularnewline
\hline
\multirow{2}{*}{2-Frame Pretraining} & No & 1.16 & \textbf{2.18} & 4.27 & 16.55\tabularnewline
 & \uline{Yes} & \textbf{1.00} & 2.19 & \textbf{3.86} & \textbf{14.76}\tabularnewline
\hline
\multirow{4}{*}{Inference Mem Length} & 0 & 1.16 & 2.41 & 4.20 & 15.44\tabularnewline
 & \uline{1} & \textbf{1.00} & \textbf{2.19} & \textbf{3.86} & 14.76\tabularnewline
 & 2 & 1.02 & 2.36 & \textbf{3.86} & 14.65\tabularnewline
 & 3 & 1.07 & 2.25 & 3.87 & \textbf{14.64}\tabularnewline
\hline
\tabularnewline
\multicolumn{6}{c}{\noindent\emph{Reference Model, Training: 600k on FlyingThings, max Mem is 2}}\tabularnewline
\hline
\multirow{3}{*}{Training Video Length} & \uline{3} & \textbf{0.93} & \textbf{2.08} & 3.88 & \textbf{13.71}\tabularnewline
 & 5 & 0.97 & 2.11 & 3.80 & 14.14\tabularnewline
 & 8 & 0.95 & \textbf{2.08} & \textbf{3.64} & 14.65\tabularnewline
\hline
\multirow{2}{*}{Warm-start} & With & 1.02 & 2.14 & 3.92 & \textbf{13.70}\tabularnewline
 & \uline{Without} & \textbf{0.93} & \textbf{2.08} & \textbf{3.88} & 13.71\tabularnewline
\bottomrule
\end{tabular}
\vspace{-0.2in}
\end{table}

\noindent\textbf{Resolution-adaptive Re-scaling.} We ablate the proposed resolution-adaptive re-scaling on the newly proposed 1080p Spring dataset, which has optical flow groundtruth at a resolution of 1920x1080 and therefore an ideal testbed for research of cross-resolution generalization. MemFlow is trained at a resolution of 368x768 by default, while we also train another model at a much higher resolution of 432x960 for comparison. As shown in \cref{tab:abla_ada_weight}, our proposed method can substantially improves the cross-resolution generalization performance. Compared to high-resolution finetuning, our method also performs better with minimal training cost.

\begin{table} \small
\centering
\caption{Resolution-adaptive re-scaling substantially improves the generalization to Full-HD (\emph{i.e.}, 1920x1080) Spring training set.\label{tab:abla_ada_weight}}
\vspace{-0.1in}
\begin{tabular}{cccccc}
\toprule
\multirow{2}{*}{Method} & \multicolumn{4}{c}{1px} & \multirow{2}{*}{EPE}\tabularnewline
\cline{2-5}
 & total & s0-10 & s10-40 & s40+ & \tabularnewline
\midrule
None & 4.245 & 2.535 & 12.216 & 42.630 & 0.448\tabularnewline
High-reso. ft & 4.456 & 2.741 & 12.360 & 43.174 & 0.436\tabularnewline
Ada. Re-scaling & \textbf{4.211} & \textbf{2.512} & \textbf{12.113} & \textbf{42.404} & \textbf{0.433}\tabularnewline
\bottomrule
\end{tabular}
\vspace{-0.2in}
\end{table}


\textbf{Discussion: why set max memory length to 1?} 
We've demonstrated that optimizing the memory buffer's maximum length to 1 achieves the best performance. Essentially, this implies that our model operates in a mode similar to a 3-frame setup. The underlying intuition behind this is that a 3-frame video represents the minimal length needed for effective temporal modeling, ensuring the most consistent motion along the time axis for the same object. Moreover, a 3-frame configuration already encompasses all the matching information required for the center frame~\cite{janai2018unsupervised}. In simpler terms, if there's a pixel in the center frame, it's usually visible in at least one other frame. This is probably one reason why various methods~\cite{janai2018unsupervised, ren2019fusion, shi2023videoflow, liu2019selflow, liu2020learning} opt for a 3-frame approach in flow estimation.
However, it's important to note that our MemFlow operates differently. Our memory module efficiently accumulates and updates the motion state without redundant computations over time and doesn't explicitly extract a 3-frame sequence for estimation.


\textbf{Discussion: why not a longer range?}
We have taken an initial step in capturing long-range motion cues for flow estimation. Specifically, we train MemFlow on an 8-frame video and enhance motion features in the memory buffer by adding relative position encoding~\cite{su2021roformer} along the time axis. Despite these efforts, setting the memory length to 1 continues to yield the best results, as indicated in \cref{tab:abla_rope}. Furthermore, we have also experimented with distilling outdated memory features into long-term memory based on historical attention scores, following a similar approach to XMem~\cite{cheng2022xmem}. However, introducing long-term memory doesn't significantly impact performance, as evidenced in \cref{tab:abla_rope}. We believe a potential avenue for future research involves exploring long-range motion history for optical flow estimation while ensuring efficiency for real-time applications.

\begin{table} \small
\setlength{\tabcolsep}{4pt}
\centering
\caption{More ablations about memory module. All models are trained on 8 frames video with relative position encoding here.\label{tab:abla_rope}}
\vspace{-0.1in}
\begin{tabular}{cccccc}
\toprule
\multirow{2}{*}{Experiment} & \multirow{2}{*}{Method} & \multicolumn{2}{c}{Sintel} & \multicolumn{2}{c}{KITTI-15}\tabularnewline
\cline{3-6} 
 &  & Clean & Final & Fl-epe & Fl-all\tabularnewline
\midrule
\multirow{3}{*}{Inference Mem Length} & \uline{1} & 1.05 & \textbf{2.12} & \textbf{3.65} & \textbf{13.78}\tabularnewline
 & 2 & 1.06 & 2.27 & \textbf{3.65} & 13.81\tabularnewline
 & 3 & \textbf{1.03} & 2.18 & 3.67 & 13.83\tabularnewline
\hline
\multirow{2}{*}{Long Term Mem} & With & 1.05 & 2.12 & 3.65 & \textbf{13.78}\tabularnewline
 & \uline{Without} & \textbf{1.03} & \textbf{2.10} & \textbf{3.64} & 13.80\tabularnewline
\bottomrule
\end{tabular}
\vspace{-0.25in}
\end{table}

\section{Conclusion}
\label{sec:Conclusion}
We introduced MemFlow, a novel online approach for video-based optical flow estimation and prediction. What sets MemFlow apart is its use of a memory module to store historical motion states. Additionally, MemFlow incorporates resolution-adaptive re-scaling, enhancing cross-resolution performance at minimal training cost. Notably, MemFlow stands out with state-of-the-art cross-dataset generalization and high inference efficiency. Besides, with minimal adjustments, MemFlow can be repurposed for flow prediction, achieving top-notch prediction performance.

{\small\noindent\textbf{Acknowledgements:} 
Yanwei Fu is the corresponding authour. Yanwei Fu 
is also with Shanghai Key Lab of Intelligent Information Processing, Fudan University, and 
Fudan  ISTBI-ZJNU Algorithm Centre 
for Brain-inspired Intelligence, 
Zhejiang Normal University.
The computations in this research were performed using the CFFF platform of Fudan University.}

{
    \small
    \bibliographystyle{ieeenat_fullname}
    \bibliography{arxiv_main}
}

\clearpage
\setcounter{page}{1}
\maketitlesupplementary

\section{Details of Future Prediction}
\label{sec:supp_future_prediction}

\begin{figure*}
\centering
\includegraphics[width=0.95\linewidth]{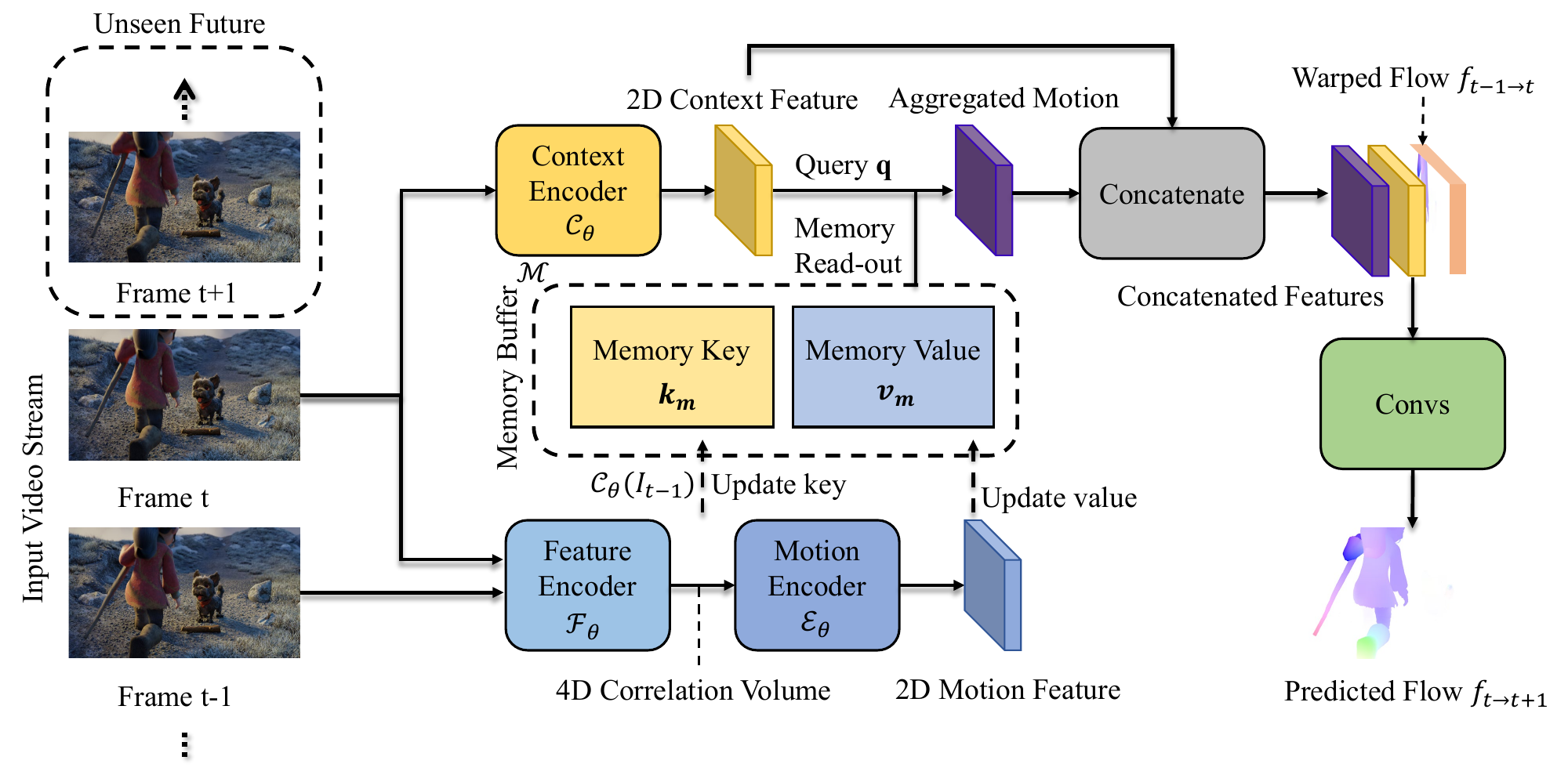}
\caption{Overview of our MemFlow-P for future prediction of optical flow.\label{fig:ofp_overview_supp}}
\end{figure*}

\noindent\textbf{MemFlow-P}. We present an overview of our Memory module for future Prediction of optical Flow (MemFlow-P) as in \cref{fig:ofp_overview_supp}. Specifically, given current frame $\mathbf{I}_t$, we should first calculate the 2D motion feature $f_{m}$ with previous frame $\mathbf{I}_{t-1}$. We are now able to update the memory buffer with $f_{m}$ and the context feature $\mathcal{C}_\theta(\mathbf{I}_{t-1})$ from $\mathbf{I}_{t-1}$. Then we extract the context feature $\mathcal{C}_\theta(\mathbf{I}_{t})$ from the current frame $\mathbf{I}_t$, which also serves as a query and reads out the aggregated motion feature $f_{am}$ from the memory buffer. Besides, we also forward warp the previous flow $f_{t-1\rightarrow t}$ as a base $f_p$ for flow prediction. Finally, we concatenate the aggregated motion feature from history, context feature from the current frame, and forward warped flow $f_p$ for flow prediction with a simple CNN: $f=\mathrm{Convs}(f_c, f_{am}, f_p)$. Our CNN has similar convolutional layers as the original GRU. It consists of two SKBlocks as introduced by SKFlow~\cite{sun2022skflow}. Each SKBlock consists of two Feed Forward Networks (FFN), two depth-wise convolutional layers, and one point-wise convolutional layer. The total parameter of our MemFlow-P is 5.1 M. Our loss function is the $l_1$ distance between our predicted flow and the groundtruth:
\begin{equation}
\mathcal{L}=||f_{gt}-f||_1.
\end{equation}

\begin{figure*}
\centering
\includegraphics[width=0.95\linewidth]{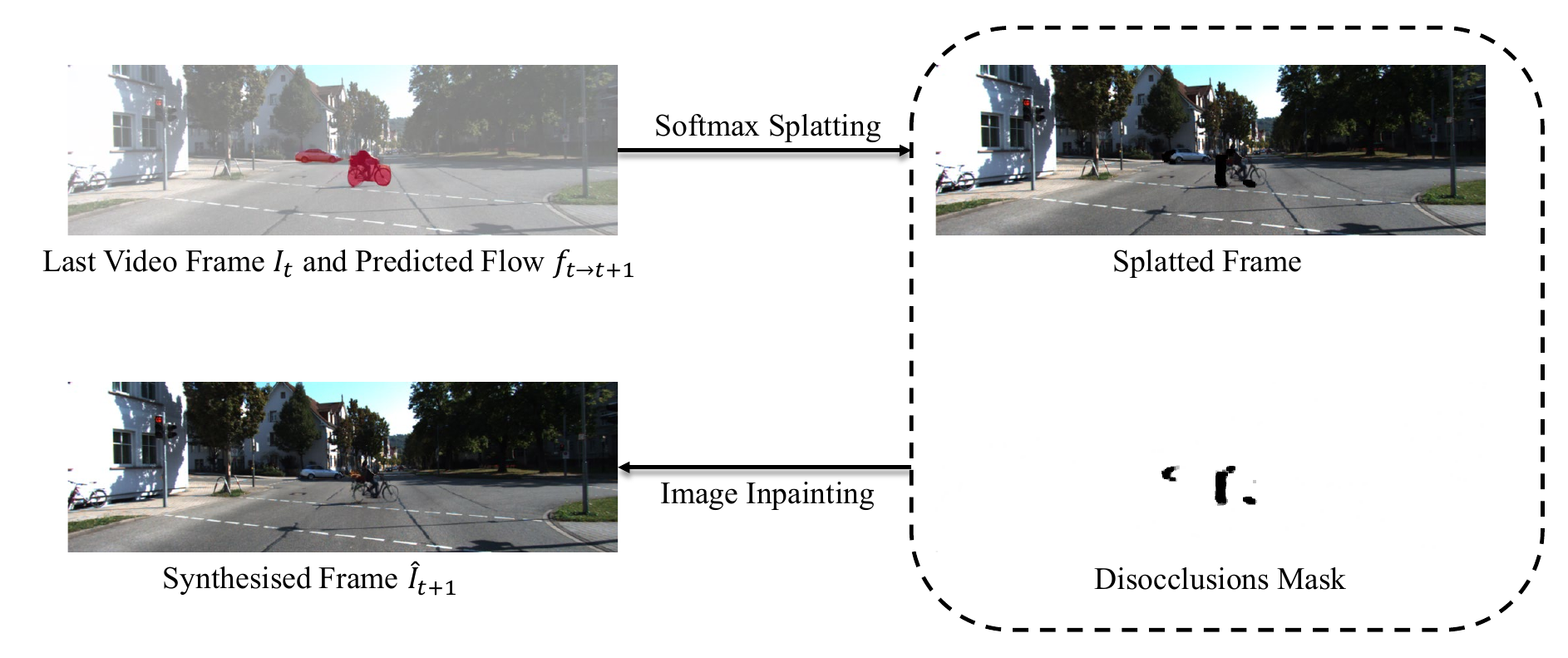}
\caption{Overview of our MemFlow-P for video prediction.\label{fig:video_prediction_supp}}
\end{figure*}

\noindent\textbf{MemFlow-P for Video Prediction}. As shown in \cref{fig:video_prediction_supp}, we first predict the optical flow $f_{t\rightarrow t+1}$ for the last video frame $\mathbf{I}_t$. Besides, we also estimate the monocular depth from DPT~\cite{ranftl2021vision} for the last video frame. We then utilize the Softmax Splatting~\cite{Niklaus_CVPR_2020} for forward warping the last video frame. As shown in the right part of \cref{fig:video_prediction_supp}, we get the splatted frame and a disocclusion mask indicating the blank regions. We finally inpaint the disocclusion region with image inpainting method ZITS~\cite{dong2022incremental} and get the synthesised frame $\hat{\mathbf{I}}_{t+1}$.

\section{Implementation Details}

\noindent\textbf{Network Details}. Our \textbf{MemFlow} shares the same network architecture with SKFlow~\cite{sun2022skflow}. Specifically, our feature encoder and context encoder consist of 6 residual blocks, 2 at 1/2 resolution, 1/4 resolution, and 1/8 resolution, respectively. Besides, our motion encoder and GRU are based on 6 and 2 SKBlocks as in SKFlow~\cite{sun2022skflow}, respectively. And our \textbf{MemFlow-P} only replaces the GRU with a small CNN as illustrated in \cref{sec:supp_future_prediction}. As for our \textbf{MemFlow-T}, we utilize the first two stages of ImageNet-pretrained Twins-SVT~\cite{Chu2021twins} as our feature and context encoder.

\noindent\textbf{Training Details}. During training, we employ Flash{A}ttention-2~\cite{dao2022flashattention,dao2023flashattention2} for faster memory read-out.

\noindent\textbf{Training Schedule}. We first pre-train our networks with 2-frame in FlyingChair and FlyingThings3D for 120k (batch size 8) and 150k (batch size 6) iterations, respectively. Then, we train our networks with 3-frame and batch size 8 on the following datasets, for
\begin{itemize}
    \item \textbf{MemFlow}, we train on FlyingThings3D for additional 600k iterations for generalization evaluation. Then, we finetune our model for 600k iterations on Sintel, KITTI, HD1K, and FlyingThings3D for Sintel submission. Finally, we finetune on KITTI for 40k and on Spring for 400k iterations, respectively.
    \item \textbf{MemFlow-T}, we train on FlyingThings3D for additional 600k iterations for generalization evaluation. Then, we finetune our model for 300k iterations on Sintel, KITTI, HD1K, and FlyingThings3D for Sintel submission. Finally, we finetune on KITTI for 40k iterations.
    \item \textbf{MemFlow-P}, we randomly initialized the newly added CNN. We then train MemFlow-P on FlyingThings3D for an additional 40k iterations for generalization evaluation. For the experiment of video prediction, we train our models on Sintel, KITTI, HD1K, and FlyingThings3D with 300k iterations.
\end{itemize}

\noindent\textbf{Evaluation Protocol of Video Prediction}. We evaluate the performance of video prediction on four sequences from the KITTI test set following previous works~\cite{geng2022comparing,wu2022optimizing}. The four sequences we employed are: 
\begin{itemize}
\item "2011\_09\_26\_drive\_0060\_sync", \item "2011\_09\_26\_drive\_0084\_sync", \item "2011\_09\_26\_drive\_0093\_sync", and
\item "2011\_09\_26\_drive\_0096\_sync".
\end{itemize}
Besides, as in prior works~\cite{geng2022comparing,wu2022optimizing}, we use a context of T=4 past frames as input. All algorithms synthesize the next frame based on past frames.

\section{More Qualitative Comparison}
\begin{figure*}
\centering
\includegraphics[width=0.95\linewidth]{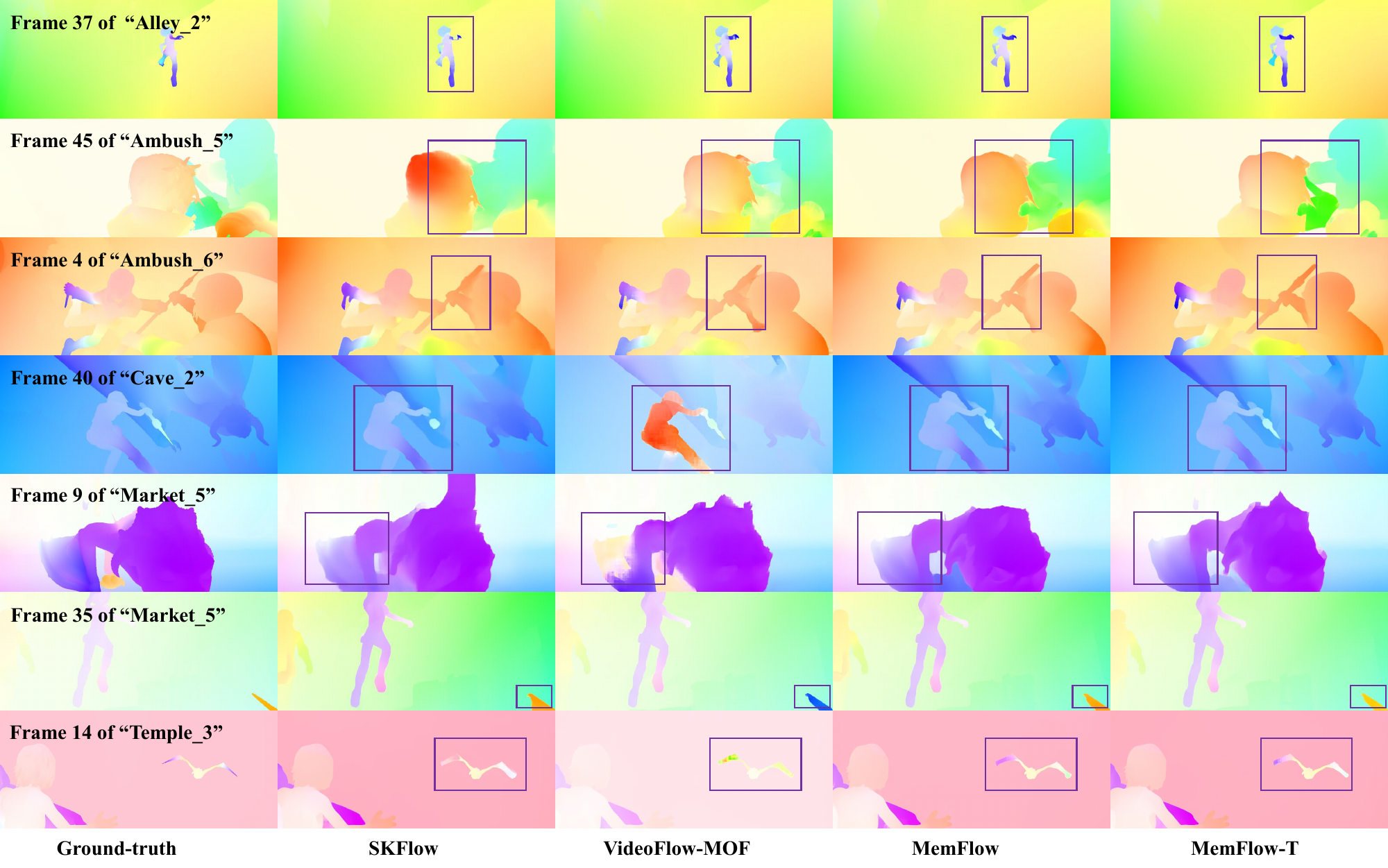}
\caption{More qualitative results on Sintel training set final pass after pre-training on FlyingChair and FlyingThings3D. Bounding boxes mark the regions of substantial improvements. Please zoom in for details.\label{fig:sintel_qualitative_supp}}
\end{figure*}

\begin{figure*}
\centering
\includegraphics[width=0.95\linewidth]{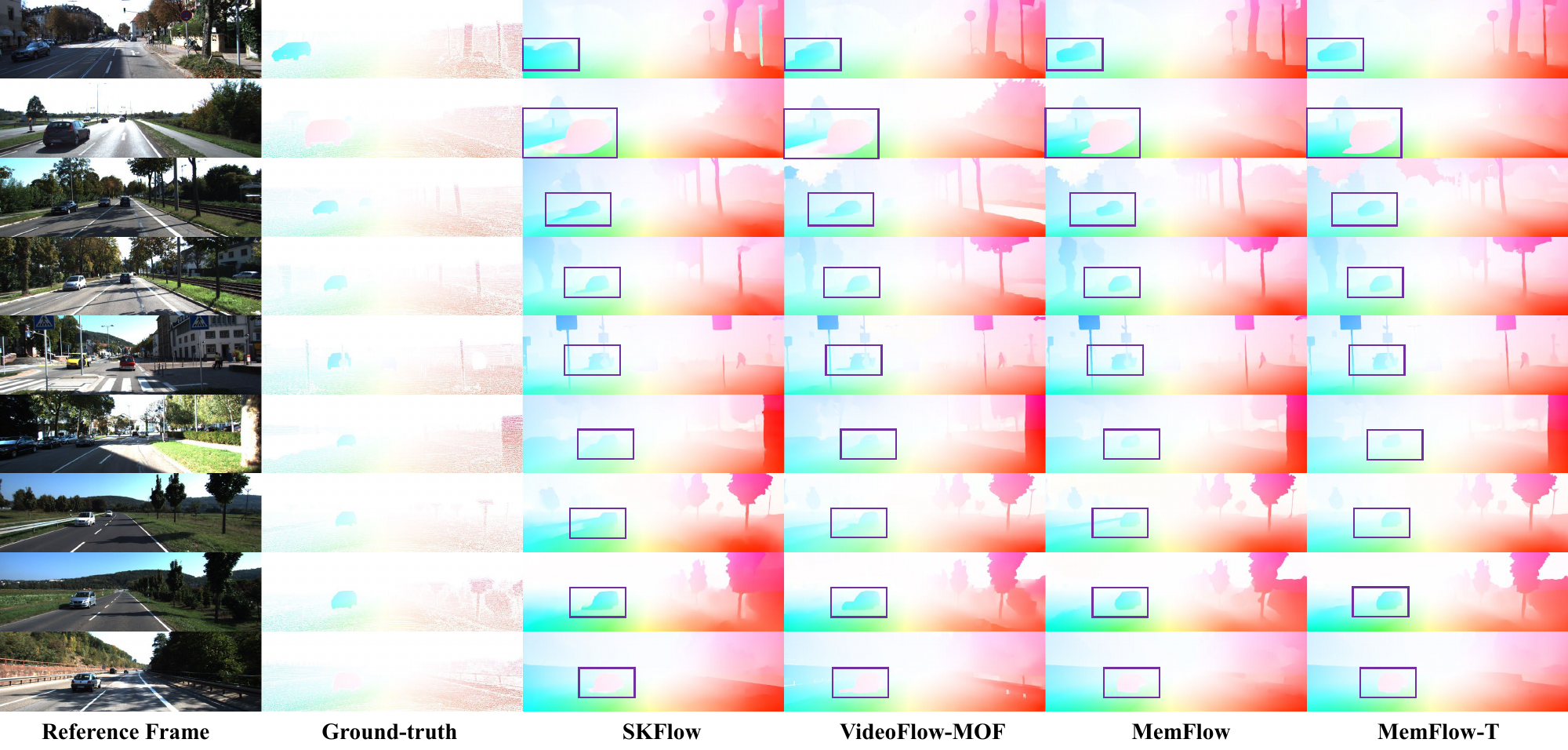}
\caption{Qualitative results on KITTI training set after pre-training on FlyingChair and FlyingThings3D. Bounding boxes mark the regions of substantial improvements. Please zoom in for details.\label{fig:kitti_qualitative_supp}}
\end{figure*}

More qualitative results on Sintel training set and KITTI training set after pre-training on FlyingChair and FlyingThings3D are given in \cref{fig:sintel_qualitative_supp,fig:kitti_qualitative_supp}. We highlight the areas where our MemFlow(-T) achieves substantial improvements with bounding boxes, compared to previous state-of-the-art VideoFlow-MOF~\cite{shi2023videoflow} and our baseline SKFlow~\cite{sun2022skflow}. Please zoom in for more details.

\begin{figure*}
\centering
\includegraphics[width=0.8\linewidth]{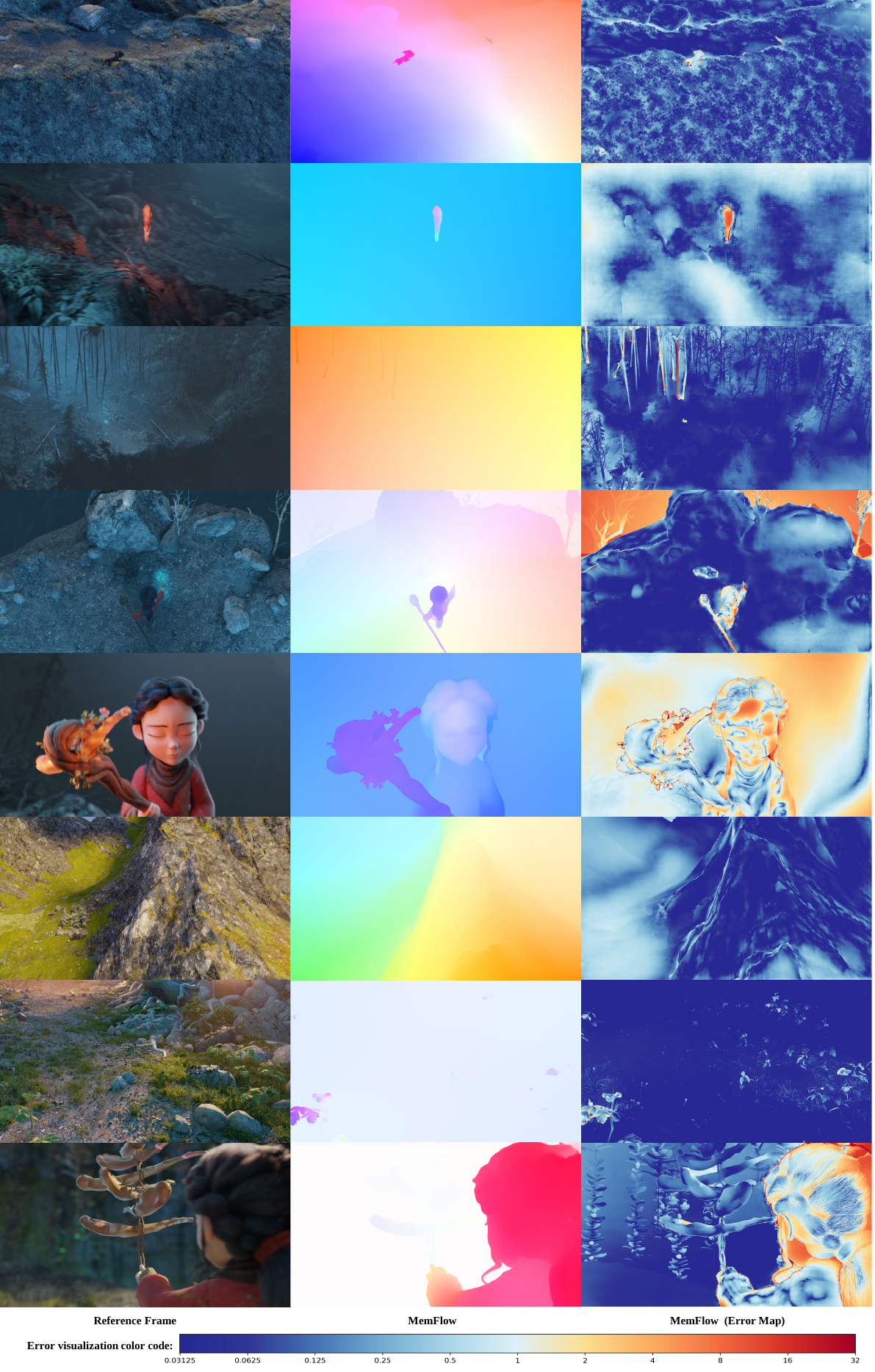}
\caption{Qualitative results on 1080p Spring test set after finetuning on Spring. Error maps are downloaded from the official website. Please zoom in for details.\label{fig:spring_qualitative_supp}}
\end{figure*}

We also provide more qualitative results on the 1080p Spring test set as shown in \cref{fig:spring_qualitative_supp}. The qualitative results show superior cross-resolution generalization performance of our MemFlow, which is trained with the image resolution of 368x768.

\section{More Results on Future Prediction of Optical Flow}
\noindent\textbf{Flow Prediction Results}. We further show the full results of generalization performance evaluation for flow prediction in \cref{tab:supp_ofp}.  MemFlow-P still outperforms other competitors in terms of the EPE on the clean pass of datasets and the Fl-all on KITTI-15 by a large margin, showing great dataset-specific and cross-dataset performance.

\begin{table} \small
\setlength{\tabcolsep}{4pt}
\centering
\caption{Generalization evaluation of flow prediction on FlyingThings3D, Sintel, and KITTI-15.\label{tab:supp_ofp}}
\begin{tabular}{ccccccc}
\toprule
\multirow{2}{*}{Method} & \multicolumn{2}{c}{Things} & \multicolumn{2}{c}{Sintel} & \multicolumn{2}{c}{KITTI-15}\tabularnewline
\cline{2-7}
 & Clean & Final & Clean & Final & Fl-epe & Fl-all\tabularnewline
\midrule
Warped Oracle & 14.76 & 14.76 & 5.76 & 5.76 & - & -\tabularnewline
\textbf{MemFlow} & 15.55 & 15.70 & 5.92 & 6.23 & 12.95 & 54.48\tabularnewline
\hdashline
OFNet~\cite{ciamarra2022forecasting} & 13.73 & 13.76 & 5.78 & 6.03 & 12.43 & 59.17\tabularnewline
\textbf{MemFlow-P} & \textbf{7.81} & \textbf{7.56} & \textbf{4.97} & \textbf{5.38} & \textbf{8.82} & \textbf{43.93}\tabularnewline
\bottomrule
\end{tabular}
\end{table}

\noindent\textbf{Ablation Studies}. In this section, we report the ablation studies of flow prediction. First, we train a baseline model for flow prediction without the forward warped past flow as input of CNN. The model is trained with 6-frame videos sampled from FlyingThings3D. As shown in \cref{tab:supp_ofp_abla}, concatenating the forward warped flow can improve the cross-dataset generalization performance a lot, though with little worse results on the FlyingThings3D test split. Moreover, we find that training MemFlow-P with 3-frame videos can achieve similar results as the one trained with 6-frame. Therefore, we choose to train our MemFlow-P with 3-frame videos and forward warped flow.

\begin{table} \small
\setlength{\tabcolsep}{3pt}
\centering
\caption{Ablation studies on optical flow prediction.\label{tab:supp_ofp_abla}}
\begin{tabular}{ccccccc}
\toprule
\multirow{2}{*}{Experiment} & \multicolumn{2}{c}{Things} & \multicolumn{2}{c}{Sintel} & \multicolumn{2}{c}{KITTI-15}\tabularnewline
\cline{2-7}
 & Clean & Final & Clean & Final & Fl-epe & Fl-all\tabularnewline
\midrule
Baseline & \textbf{7.62} & \textbf{6.58} & 5.25 & 5.79 & 8.84 & 56.63\tabularnewline
+Forward Warped Flow & 7.76 & 7.57 & \textbf{4.96} & 5.47 & \textbf{8.57} & 53.15\tabularnewline
+Training with 3-frame & 7.81 & 7.56 & 4.97 & \textbf{5.38} & 8.82 & \textbf{43.93}\tabularnewline
\bottomrule
\end{tabular}
\end{table}

\begin{figure*}
\centering
\includegraphics[width=0.95\linewidth]{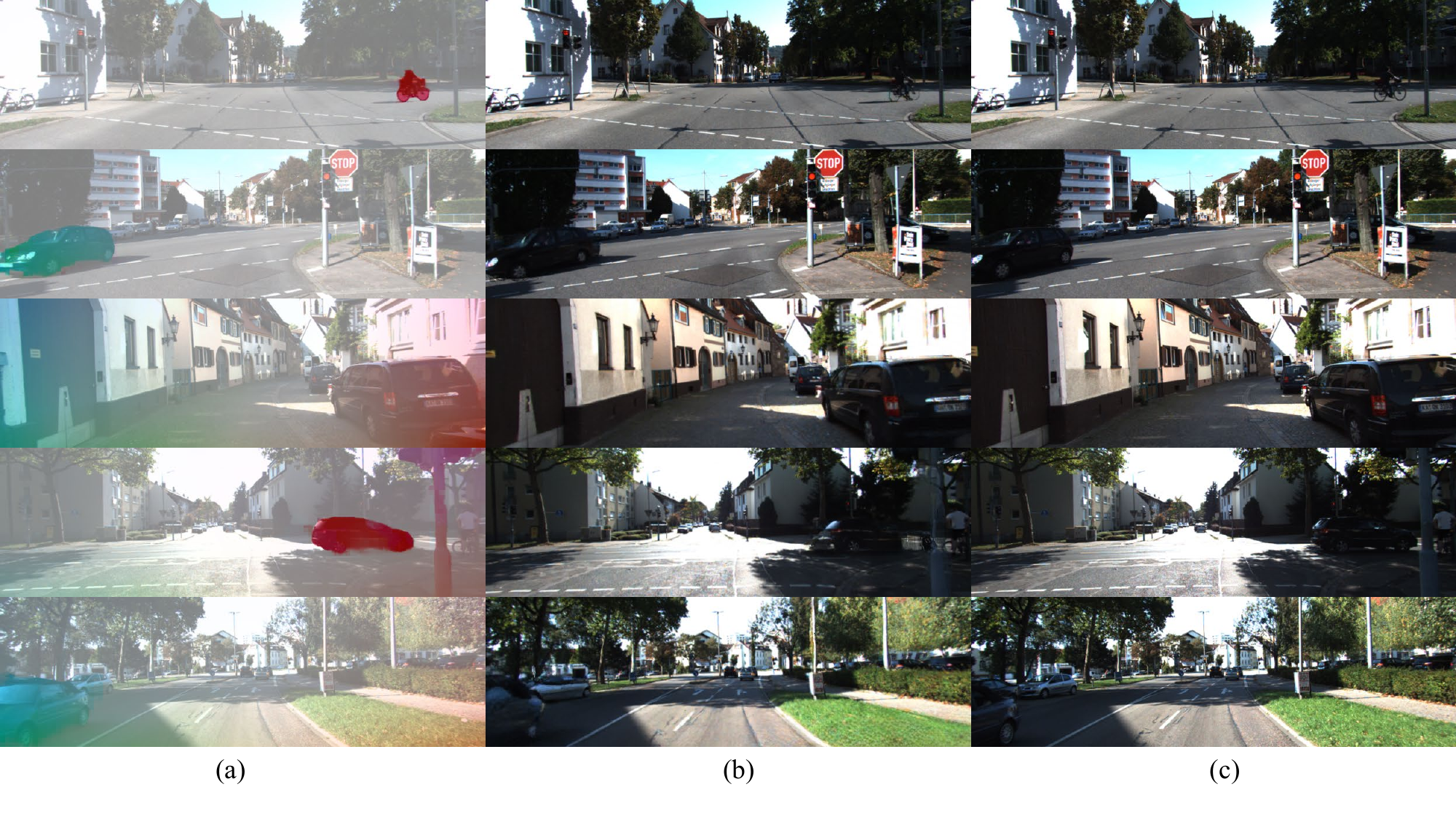}
\caption{Qualitative Results of Future Prediction by Optical Flow. (a) Predicted optical flow superimposed on the last video frame. (b) Synthesized video frame based on our predicted flow. (c) Groundtruth next frame.\label{fig:supp_ofp_qualitative}}
\end{figure*}

\noindent\textbf{Qualitative Results of Future Prediction by Optical Flow}. We further provide several qualitative results of future prediction by optical flow as shown in \cref{fig:supp_ofp_qualitative}. Our MemFlow-P can predict credible flow for the last video frame, and successfully synthesize the next frame.

\begin{figure}
\centering
\includegraphics[width=0.95\linewidth]{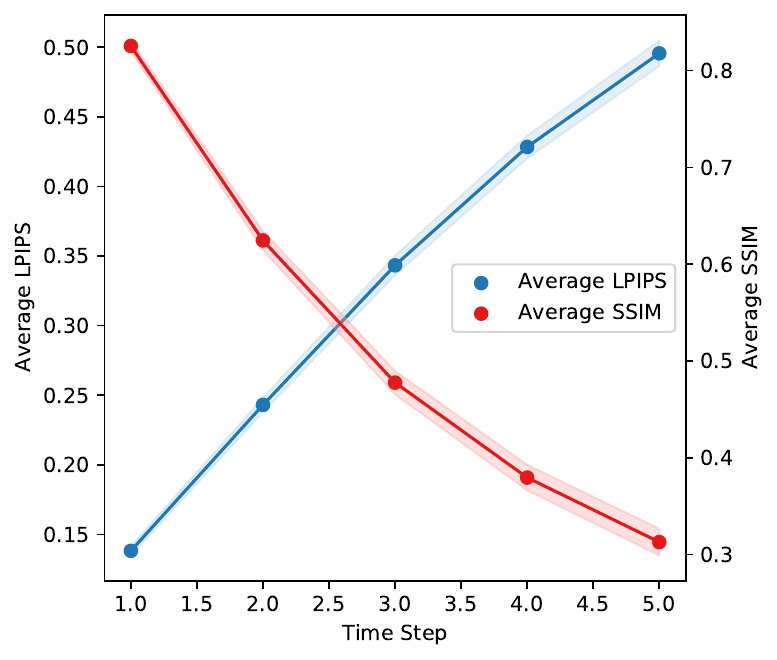}
\caption{Quantitative results of long-term future prediction by optical flow. The plot shows the average LPIPS and SSIM-time step chart over KITTI test videos (256x832) and shadow is the 95\% confidence interval. We calculate the metric with predicted frames for up to time step T + 5 from a context of T=4 past frames.\label{fig:supp_lpips_ssim_time_step_graph}}
\end{figure}

\begin{figure*}
\centering
\includegraphics[width=0.95\linewidth]{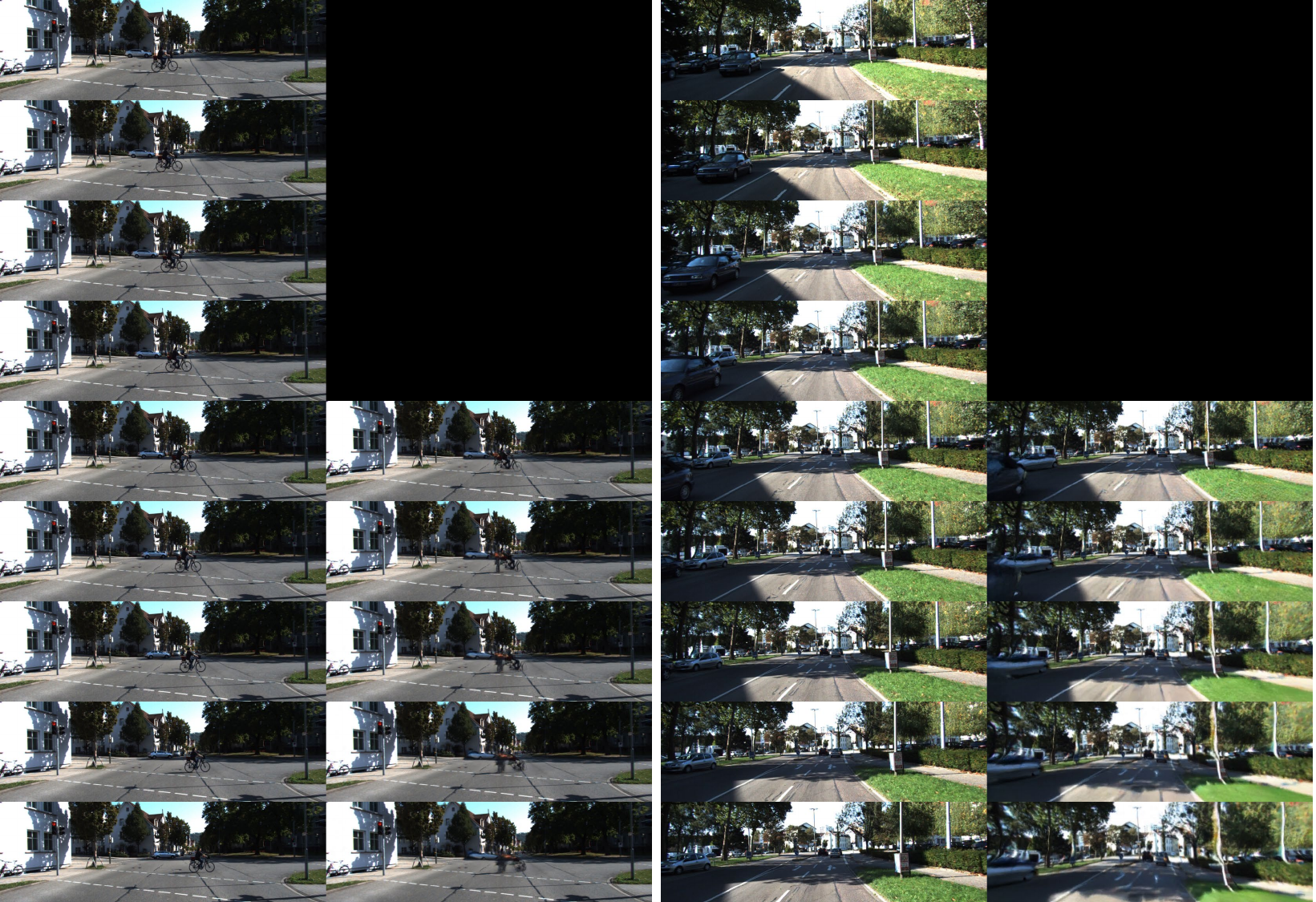}
\caption{Our approach may fail to generate high-quality frames many steps into the future autoregressively due to error accumulation: Given 4 conditioning frames (top left), we show 5 predicted future frames in column 2 (bottom right) of two videos. Groundtruth frames are shown in the bottom left. \label{fig:supp_long_term_vp_qualitative}}
\end{figure*}

\noindent\textbf{Limitations of Long-term Future Prediction by Optical Flow}. Our approach can generate nice results for short-term (one time step) future prediction as shown in \cref{fig:supp_ofp_qualitative}. However, in the long term, as the predicted frame deviates from the distribution of training images, performance will drop quickly due to error accumulation like other video prediction methods. We further provide the quantitative and qualitative results of long-term future prediction in \cref{fig:supp_lpips_ssim_time_step_graph,fig:supp_long_term_vp_qualitative}.

\section{Screenshots of 1080p Spring, Sintel, and KITTI Results}

\begin{figure*}
\centering
\includegraphics[width=0.95\linewidth]{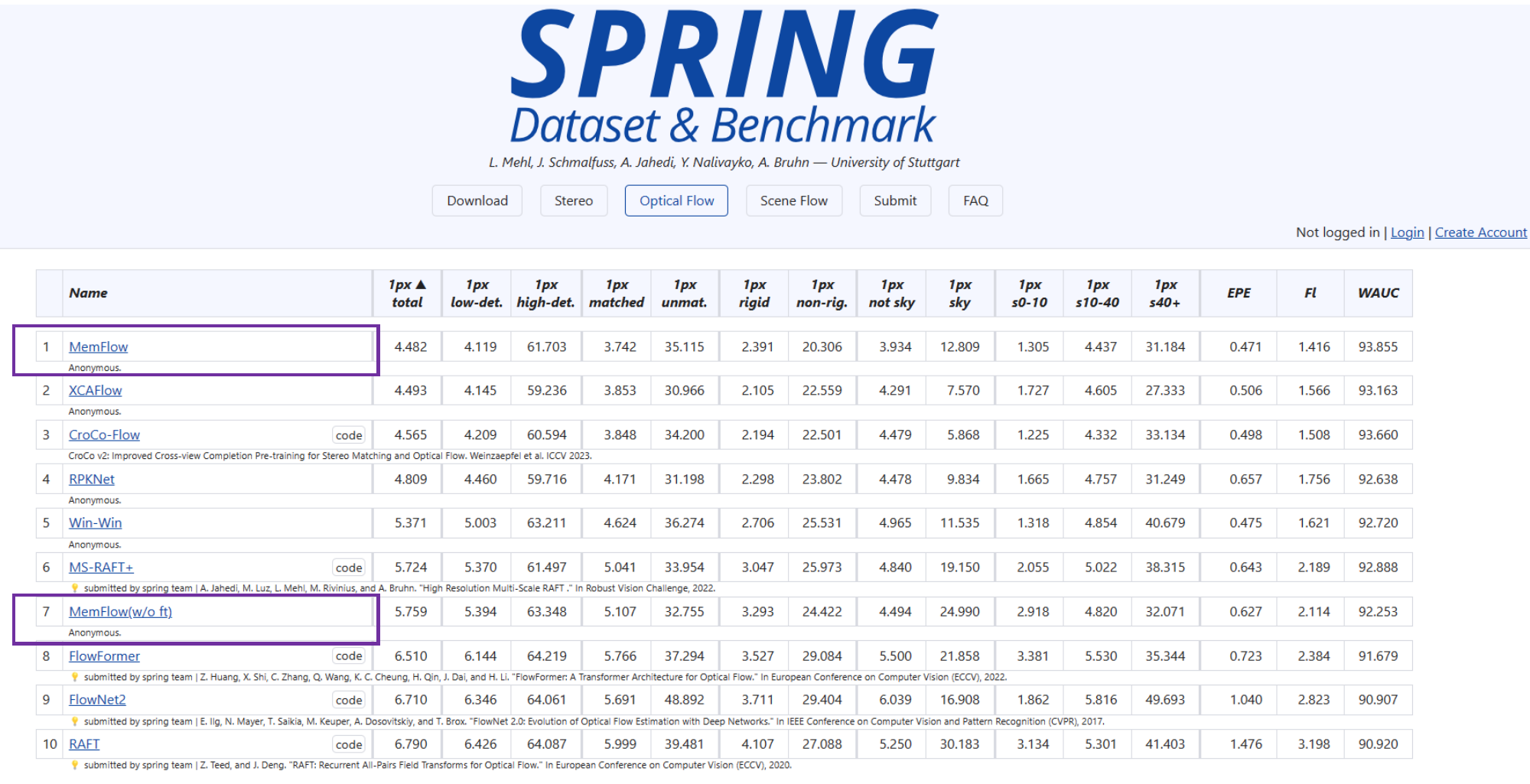}
\caption{Screenshots for 1080p Spring optical flow evaluation on the official website.\label{fig:spring_screen_shots_supp}}
\end{figure*}

\begin{figure*}
\centering
\includegraphics[width=0.95\linewidth]{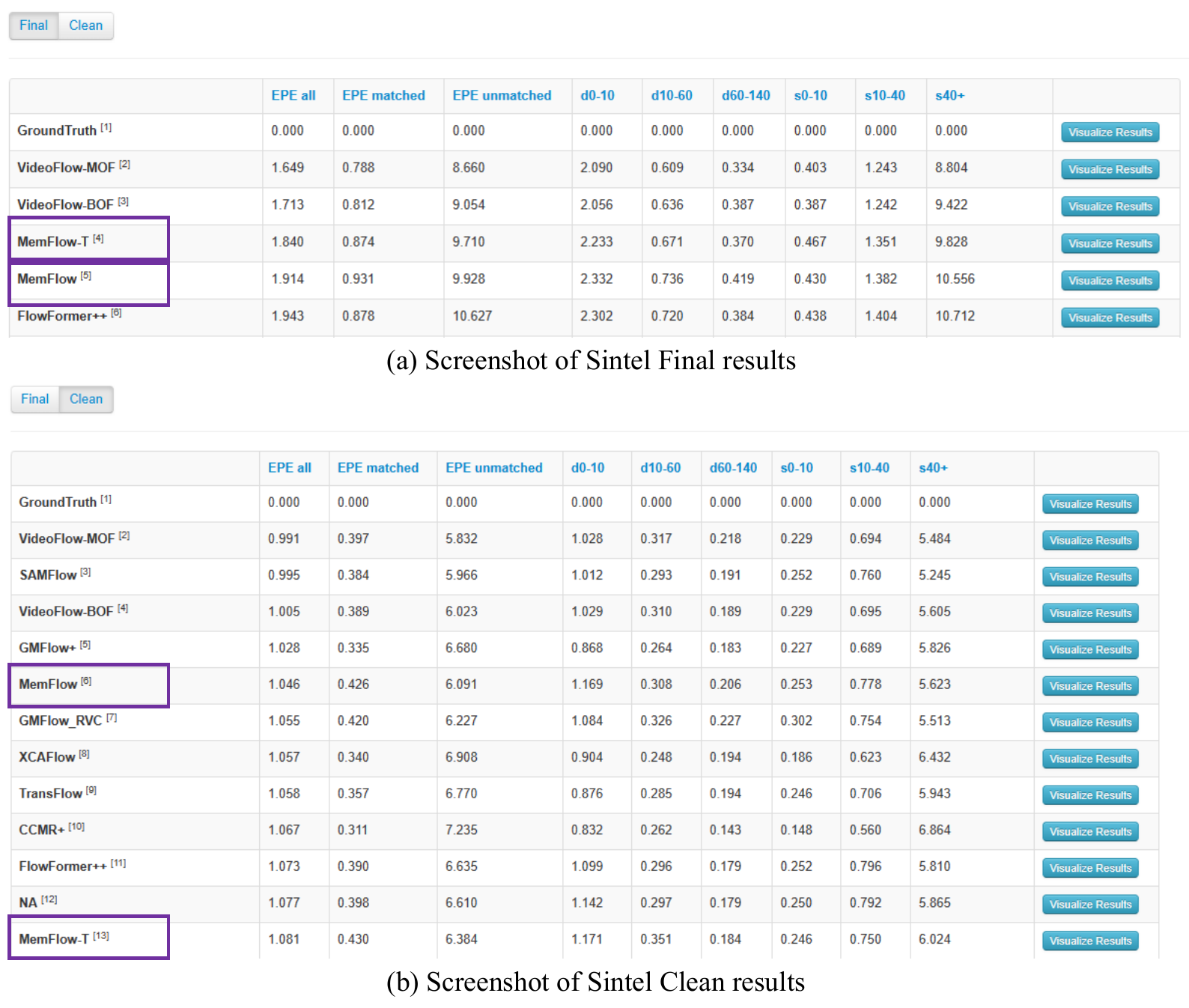}
\caption{Screenshots for Sintel optical flow evaluation on the official website.\label{fig:sintel_screen_shots_supp}}
\end{figure*}

\begin{figure*}
\centering
\includegraphics[width=0.95\linewidth]{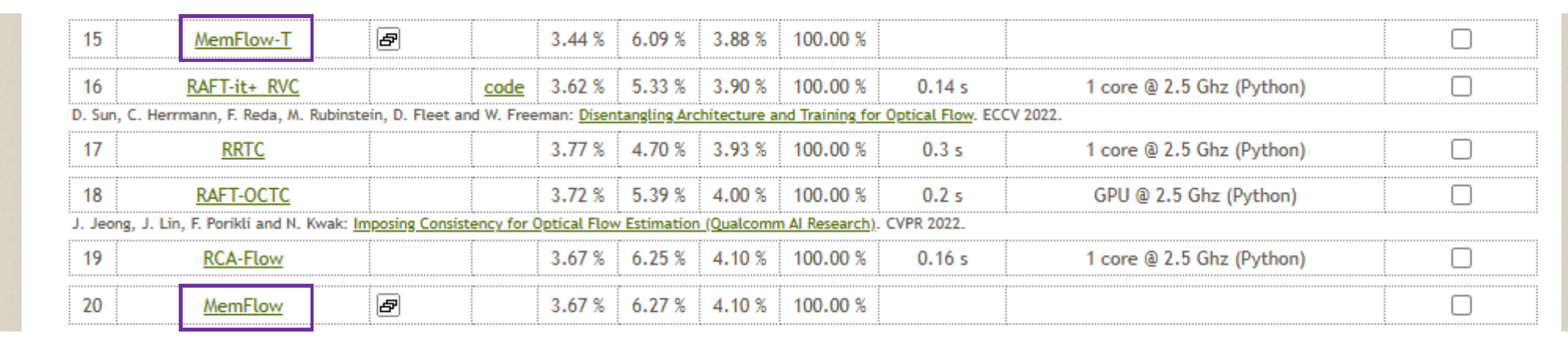}
\caption{Screenshots for KITTI-15 optical flow evaluation on the official website.\label{fig:kitti_screen_shots_supp}}
\end{figure*}

We further provide anonymous screenshots of Spring, Sintel, and KITTI results on the test server as in \cref{fig:spring_screen_shots_supp,fig:sintel_screen_shots_supp,fig:kitti_screen_shots_supp}. Our MemFlow ranks first on the 1080p Spring benchmark. The one without finetuning on Spring also performs well in terms of cross-dataset generalization performance. On Sintel, our MemFlow-T and MemFlow take the third and fourth places on the final pass, which improves the performance of SKFlow a lot. We also achieved great improvement on the KITTI benchmark compared to the baseline SKFlow.

\end{document}